\documentclass[11pt]{article}

\usepackage[preprint]{acl}
\usepackage{times}
\usepackage{latexsym}
\usepackage{makecell}
\usepackage{multirow}
\usepackage[T1]{fontenc}

\usepackage[utf8]{inputenc}
\usepackage{fancyvrb}
\usepackage{fvextra}
\usepackage{pifont}
\usepackage{graphicx}
\newcommand{\cmark}{\ding{51}}
\newcommand{\xmark}{\ding{55}}
\usepackage{microtype}
\usepackage{textgreek}
\usepackage{newunicodechar}
\newunicodechar{₂}{$_2$}
\newunicodechar{₁}{$_1$}
\newunicodechar{；}{;}
\usepackage{inconsolata}
\usepackage{graphicx}
\usepackage{fancyvrb}
\usepackage{pgfplots}
\usepackage[table]{xcolor}
\definecolor{cvprblue}{rgb}{0.21,0.49,0.74}

\usepackage{array}
\usepackage{booktabs}
\usepackage{siunitx}
\usepackage{tabularx}
\usepackage{tikz}
\usepackage{graphicx}
\usepackage{amssymb}
\usepackage[most]{tcolorbox}
\usepackage{fancyvrb}
\newcolumntype{C}[1]{>{\centering\arraybackslash}m{#1}}
\newcolumntype{L}[1]{>{\raggedright\arraybackslash}m{#1}}
\newcolumntype{Y}{>{\raggedright\arraybackslash}X}

\definecolor{fund}{HTML}{DDEBFF} 
\definecolor{stru}{HTML}{DDF3E5} 
\definecolor{proc}{HTML}{EADCF9} 
\definecolor{func}{HTML}{FFE0E0} 

\definecolor{hm1}{HTML}{F8766D}
\definecolor{hm2}{HTML}{F4A582}
\definecolor{hm3}{HTML}{FDDC8A}
\definecolor{hm4}{HTML}{D9EF8B}
\definecolor{hm5}{HTML}{A6D96A}
\definecolor{hm6}{HTML}{66BD63}

\definecolor{tabrule}{HTML}{B8B8B8}

\newcommand{\ScoreMin}{0.021}
\newcommand{\ScoreMax}{0.557}

\newcommand{\scorecell}[1]{%
  \pgfmathtruncatemacro{\ScoreBin}{%
    min(5,max(0,floor(6*(#1-\ScoreMin)/(\ScoreMax-\ScoreMin))))%
  }%
  \ifcase\ScoreBin
    \cellcolor{hm1!40}%
  \or
    \cellcolor{hm2!40}%
  \or
    \cellcolor{hm3!40}%
  \or
    \cellcolor{hm4!40}%
  \or
    \cellcolor{hm5!40}%
  \or
    \cellcolor{hm6!40}%
  \fi
  \num{#1}%
}

\newcommand{\avgcell}[1]{%
  {\bfseries\num{#1}}%
}
\newcommand{\heatmaplegend}{%
\begin{center}
\scriptsize
\begin{tikzpicture}[baseline=-0.5ex]
  \node[anchor=east,font=\bfseries] at (0,0.09) {Worst};
  \fill[hm1!86] (0.10,0) rectangle +(0.36,0.18);
  \fill[hm2!86] (0.46,0) rectangle +(0.36,0.18);
  \fill[hm3!86] (0.82,0) rectangle +(0.36,0.18);
  \fill[hm4!86] (1.18,0) rectangle +(0.36,0.18);
  \fill[hm5!86] (1.54,0) rectangle +(0.36,0.18);
  \fill[hm6!86] (1.90,0) rectangle +(0.36,0.18);
  \draw[black!35, line width=0.2pt] (0.10,0) rectangle (2.26,0.18);
  \node[anchor=west,font=\bfseries] at (2.36,0.09) {Best};
\end{tikzpicture}
\end{center}
}
\title{\textsc{OmniMatBench}: A Human-Calibrated Multimodal Reasoning Benchmark Across 19 Materials Science Subfields}

\author{First Author \\
  Affiliation / Address line 1 \\
  Affiliation / Address line 2 \\
  Affiliation / Address line 3 \\
  \texttt{email@domain} \\\And
  Second Author \\
  Affiliation / Address line 1 \\
  Affiliation / Address line 2 \\
  Affiliation / Address line 3 \\
  \texttt{email@domain} \\}

\author{
  \textbf{Wanhao Liu\textsuperscript{1,2*}},
  \textbf{Jiaqing Xie\textsuperscript{2,3*}},
  \textbf{Qian Tan\textsuperscript{1,2}},
  \textbf{Weida Wang\textsuperscript{2,3}},
  \textbf{Jue Wang\textsuperscript{1}},
  \textbf{Ran Sun\textsuperscript{1}},
  \textbf{Zhuo Yang\textsuperscript{4}}\\
  \textbf{Wanli Ouyang\textsuperscript{2}},
  \textbf{Lei Bai\textsuperscript{2}},
  \textbf{Tianfan Fu\textsuperscript{5,2}},
  \textbf{Lu Chen\textsuperscript{6,7}},
  \textbf{Xin Chen\textsuperscript{6}$^\dagger$},
  \textbf{Yuqiang Li\textsuperscript{2}$^\dagger$}
\\
\\
  \textsuperscript{1}University of Science and Technology of China, 
  \textsuperscript{2}Shanghai Artificial Intelligence Laboratory, \\
  \textsuperscript{3}Fudan University, 
  \textsuperscript{4}Southeast University, 
  \textsuperscript{5}Nanjing University,\\
  \textsuperscript{6}Suzhou Laboratory
  \textsuperscript{7}Shanghai Jiao Tong University
\\
  \small \{liuwanhao, xiejiaqing,liyuqiang\}@pjlab.org.cn}

\begin{document}
\maketitle
\begin{abstract}

As multimodal language models play an increasingly important role in scientific research, materials science offers a critical testbed due to its interdisciplinary, multimodal, and application-driven nature. 
However, existing materials benchmarks mainly focus on property prediction, knowledge QA, or characterization understanding, leaving the broader reasoning process from materials knowledge to application underexplored. 
To fill this gap, we present \textit{OmniMatBench}, a human-calibrated multimodal reasoning benchmark for materials science. 
OmniMatBench contains 3,171 expert-curated QA and calculation problems across 19 materials-science subfields, spanning fundamental materials knowledge, structural and engineering materials, materials processing and manufacturing, and functional and applied materials. 
We evaluate 13 open-source and closed-source MLLMs and find that the best model achieves only a 0.372 overall score, revealing a substantial gap in current materials-science reasoning. 
Further analysis shows strong variation across subfields, fixed reasoning heuristics, uneven materials knowledge, and limited high-level knowledge application under formula-, retrieval-, and code-assisted settings. 
OmniMatBench provides crucial insights into the capabilities and limitations of current MLLMs and establishes a foundation for reliable AI assistants in materials-science research.

\end{abstract}

\section{Introduction}
Multimodal large language models (MLLMs) are increasingly expected to assist scientific research~\cite{tang2026multimodal, cui20252m}, yet their reliability in domain-specific scientific reasoning remains unclear. Materials science provides a particularly challenging and valuable testbed because it is inherently interdisciplinary, multimodal, and engineering-application ~\cite{lendvai2024rise, miret2025enabling}.
However, existing materials-science benchmarks largely focus on property prediction~\cite{dunn2020benchmarking, tran2023open}, knowledge QA~\cite{weng2025matqna, wu2025seeing, zaki2024mascqa}, visual recognition~\cite{cheung2025msqa}, characterization understanding~\cite{alampara2025probing}, or final-answer accuracy~\cite{zhang2025matscibench}. However, existing protocols lack systematic evaluation of materials-science contexts spanning knowledge understanding, structure--property interpretation, processing, fabrication, and application decision making.



To address this gap, we introduce \textsc{OmniMatBench}, a human-calibrated multimodal reasoning benchmark for materials science. It contains 3,171 expert-curated QA and calculation problems across 19 subfields, organized into four broad domains: \textit{Fundamental Materials Science}, \textit{Structural and Engineering Materials}, \textit{Materials Processing and Manufacturing}, and \textit{Functional and Applied Materials}. Each problem is paired with a proof record documenting its source, formulation, and solution logic, and is verified by materials-domain experts for correctness and answer consistency. This taxonomy reflects the \textit{Knowledge--Structure--Processing--Application} (KSPA) perspective, covering both foundational and engineering-oriented topics, including hard-to-collect and important subfields such as powder materials, gem materials and gemology,  welding technology and engineering, and nanomaterials and technology.

\begin{figure*}[!htbp]
    \centering
    \includegraphics[width=0.86\linewidth]{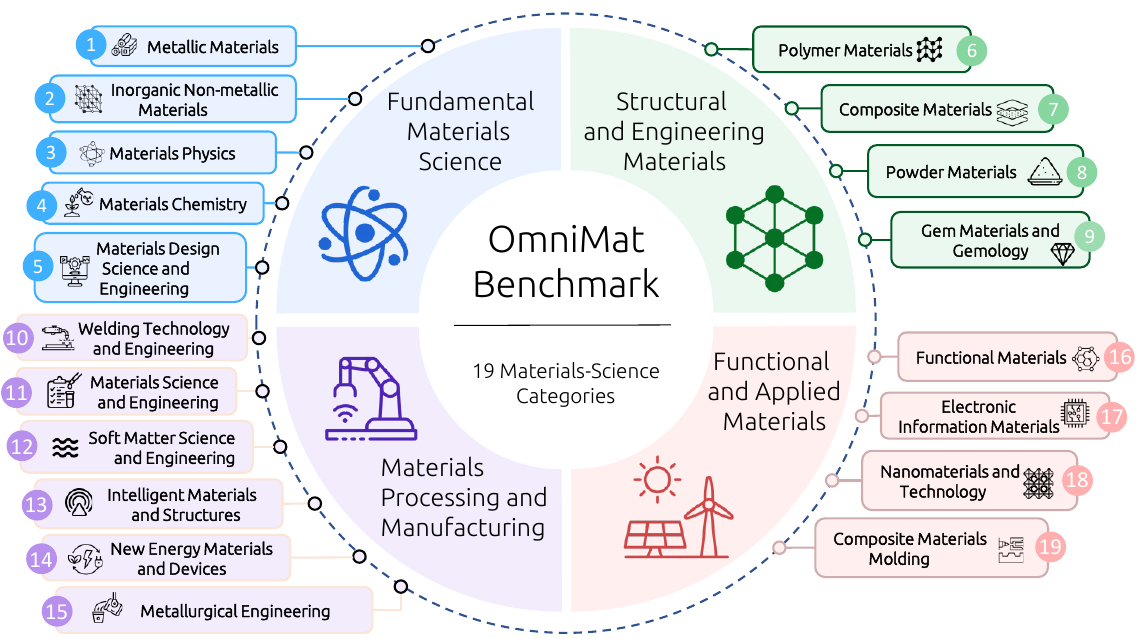}
    \caption{Overview of the OmniMatBench, illustrating the classification of 19 distinct materials-science subfields into four overarching domains.}
    \label{fig:benchmark_overview}
\end{figure*}

\textsc{OmniMatBench} adopts a fine-grained evaluation design. QA items are annotated with expert-curated key points covering concepts, reasoning elements, scoring criteria, and common errors, while calculation (CAL) items include structured answer slots, formula traces, unit requirements, numerical precision constraints, and output schemas. This enables evaluation beyond final-answer matching and reveals failures in concept identification, formula selection, multimodal grounding, unit-aware computation, and answer formatting.

We systematically evaluate 13 proprietary and open-source MLLMs on \textsc{OmniMatBench}. 
Results show that current MLLMs remain far from reliable materials-science reasoning: the strongest model, Claude Opus 4.7, achieves only 0.372 overall, with 0.443 QA F$_1$ and 30.10\% CAL slot accuracy, while GPT-5.5 reaches 0.324 overall. 
Further analysis reveals strong subfield variation, fixed reasoning heuristics, uneven materials knowledge, and persistent execution failures even with formula, retrieval, and code assistance. 
Scientific-domain specialized models can outperform general-purpose models in some subfields, but a broader knowledge-to-execution gap remains.

Our contributions are summarized as follows:
\begin{itemize}
    \item We introduce \textsc{OmniMatBench}, a broad-coverage materials-science benchmark with 3,171 expert-curated QA and CAL problems across 19 subfields and 4 domains, covering foundational and engineering-oriented topics. 
    \item We design a fine-grained evaluation protocol combining rubric-based QA with strict slot-based CAL evaluation, supported by expert key points, common-error annotations, formula traces, unit requirements, numerical precision constraints, and output schemas.
    
    \item We evaluate 13 MLLMs on \textsc{OmniMatBench} and analyze their reasoning behaviors, revealing persistent gaps including fixed reasoning heuristics, uneven subfield knowledge, weak processing/application understanding, and limited knowledge application.
\end{itemize}

\section{Related Works}

\paragraph{Materials Benchmarks and Evaluation Protocols.}
Materials benchmarks have evolved from predictive modeling to language-based scientific reasoning. Early benchmarks such as MatBench~\cite{dunn2020benchmarking}, Matbench Discovery~\cite{riebesell2025framework}, OC20/OC22~\cite{chanussot2021open,tran2023open}, and JARVIS-Leaderboard~\cite{choudhary2024jarvis} focus on property prediction, stability screening, atomistic simulation, and materials discovery. Recent LLM and MLLM benchmarks, including MaScQA~\cite{zaki2024mascqa}, MSQA~\cite{cheung2025msqa}, MatSciBench~\cite{zhang2025matscibench}, MaCBench~\cite{alampara2025probing}, MatVQA~\cite{wu2025seeing}, MatQnA~\cite{weng2025matqna}, MatCha~\cite{lai-etal-2025-multimodal}, and MATRIX~\cite{mcgrath2026matrix}, further evaluate materials QA, multimodal reasoning, and characterization understanding. However, most existing protocols emphasize knowledge recall, visual recognition, or final-answer accuracy, offering limited insight into formula selection, unit-aware computation, multi-slot answers, and format-constrained execution. To address these issues, OmniMatBench complements prior work by evaluating the knowledge-to-execution gap with structured slots, formula traces, retrieval settings, and fine-grained error analysis.

\begin{table}[!htbp]
\centering
\scriptsize
\setlength{\tabcolsep}{2.3pt}
\caption{
Comparison with materials-science benchmarks. \textbf{Mod.} denotes modality: \textbf{MM} for multimodal and \textbf{Text} for text-only.
\textbf{Numeric} denotes free-form numerical or calculation-oriented outputs rather than numerical options in multiple-choice questions.
\textbf{Diff.} is assigned by the best reported model score at the time of each benchmark release:
Easy ($>65$), Medium ($50$--$65$), and Difficult ($<50$).
\textbf{KSPA} denotes whether the benchmark includes subfields spanning the Knowledge--Structure--Processing--Application perspective.
}
\label{tab:benchmark_comparison}
\resizebox{\linewidth}{!}{
\begin{tabular}{lccclcc}
\toprule
\textbf{Benchmark} &
\textbf{\#Items} &
\textbf{Mod.} &
\textbf{Num.} &
\textbf{Task} &
\textbf{Diff.} &
\textbf{KSPA} \\
\midrule
MATRIX      & 470   & MM   & \xmark & Open QA       & Medium & \xmark \\
MatSciBench & 1,340 & MM   & \cmark & QA, Cal       & Easy   & \xmark \\
MaCBench    & 1,153 & MM   & \cmark & MCQ, Open QA  & Medium & \xmark \\
MSQA        & 1,757 & Text & \xmark & QA, T/F       & Easy   & \xmark \\
MatVQA      & 1,325 & MM   & \xmark & MCQ           & Medium & \xmark \\
MatQnA      & 4,968 & MM   & \xmark & MCQ, Subj. QA & Easy   & \xmark \\
MatCha      & 1,500 & MM   & \xmark & MCQ           & Medium & \xmark \\
\textbf{OmniMatBench}
            & \textbf{3,171}
                    & \textbf{MM}
                           & \cmark
                                   & \textbf{QA, Cal}
                                            & \textbf{Difficult}
                                                     & \cmark \\
\bottomrule
\end{tabular}
}
\vspace{-1mm}
\end{table}

\section{Dataset and Tasks}
\subsection{Data Collection and Processing}

\begin{figure}[h]
    \centering
    \begin{minipage}[t]{\linewidth}
        \centering
        \includegraphics[width=\linewidth]{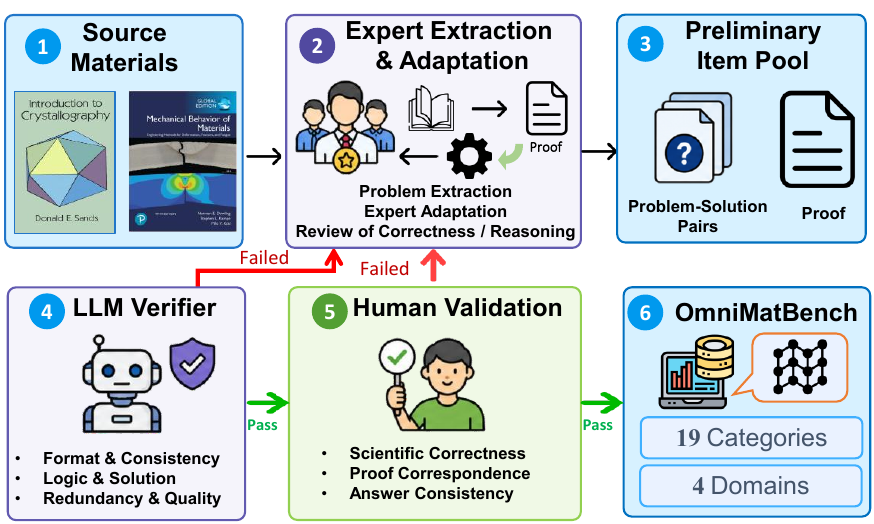}
    \end{minipage}

    \caption{
    Overview of the OmniMatBench data construction pipeline.
    }
    \label{fig:data_construction}
\end{figure}
OmniMatBench is constructed through a multi-stage, expert-involved data construction pipeline in collaboration with materials-science specialists. As shown in Figure~\ref{fig:data_construction}, the source materials are selected from classical and used materials knowledge resources, and the resulting problems are organized to cover 19 subfields aligned with the Knowledge--Structure--Processing--Application perspective.


Experts first select, refine, and complete the domain knowledge from the source materials, and then classify each candidate problem into its corresponding subfield. For each problem, they identify the core formulas, common error types, and key concepts based on their domain expertise and experience, and distill them into scoring-oriented key points. Meanwhile, the complete problem provenance, adaptation rationale, and solution logic are organized into the corresponding proof file. The preliminary item pool and proof files are then checked by an LLM verifier to ensure consistency between the question and its visual, tabular, or formula inputs, agreement between the solution and the proof file, formatting correctness, redundancy control, and overall item quality. Items that pass this stage are further reviewed by materials-domain experts, who verify disciplinary relevance, formula correctness, reasoning validity, and answer consistency. Only samples that pass both automatic verification and expert validation are included in the final OmniMatBench.





\subsection{Dataset Statistics}

OmniMatBench contains 3,171 expert-curated questions across 19 materials-science categories, spanning four major groups: \textit{Fundamental Materials Science} (FMS), \textit{Structural and Engineering Materials} (SEM), \textit{Materials Processing and Manufacturing} (MPM), and \textit{Functional and Applied Materials} (FAM). 
To address the limited coverage of specialized engineering topics in existing benchmarks~\cite{weng2025matqna, zhang2025matscibench, wu2025seeing}, OmniMatBench includes underrepresented but engineering-critical subfields, such as gem materials and gemology, powder materials, welding technology and engineering, and nanomaterials and technology. 
These areas often require specialized domain knowledge, engineering experience, and expert-verified solution procedures, making them difficult to curate at scale. 
We therefore collaborate with materials-domain experts to systematically collect and supplement representative problems from these important subfields.

The benchmark includes 1,261 open-ended QA questions and 1,910 calculation (CAL) questions. 
QA items are annotated with expert-curated key points covering core concepts, expected reasoning elements, and scoring criteria for rubric-based evaluation. 
CAL items include structured answer slots for evaluating numerical answers, formulas, units, and output formatting. 
Detailed category statistics are provided in Appendix Table~\ref{app:data-statistics}.

\paragraph{Annotation team.}
The benchmark was constructed by a team of 8 doctoral researchers, 60 trained annotators, and 5 materials-domain professors, responsible for domain-level verification, formula validation, and ambiguity resolution.

\paragraph{Difficulty distribution.}
\begin{figure}[!htbp]
    \centering
    \includegraphics[width=0.95\linewidth]{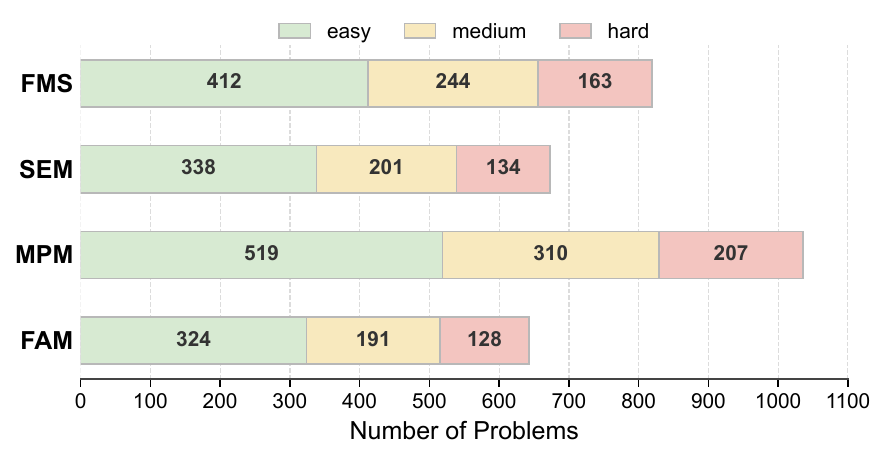}
    \caption{Expert-assigned difficulty distribution across the four major OmniMat groups.}
    \label{fig:difficulty-distribution}
\end{figure}
Each problem is assigned an expert-labeled difficulty level: easy, medium, or hard. 
Experts determine the difficulty based on the required domain knowledge, reasoning depth, multimodal interpretation, and calculation complexity. 
Figure~\ref{fig:difficulty-distribution} summarizes the difficulty distribution across the four major OmniMatBench groups, showing broadly consistent proportions despite differences in group size. 
This design supports difficulty-balanced evaluation across diverse materials-science domains.

\section{Experiments and Evaluations}
\subsection{Evaluation Setup}
\paragraph{Evaluated Models.}
We evaluate 13 frontier MLLMs, including both 8 closed-source and 5 open-source systems. 
The closed-source models include Claude Opus 4.7~\cite{anthropic2026claudeopus47}, Gemini 3.1 Pro~\cite{deepmind2026gemini31pro}, GPT-5.5~\cite{openai2026gpt55api}, Qwen3.6 Plus~\cite{qwen2026qwen36plus}, Grok-4.3~\cite{xai2026grok43}, o3, o4-mini~\cite{openai2025o3o4mini}, and GPT-4o~\cite{hurst2024gpt}, while the open-source models include Gemma-4-31B~\cite{google2026gemma4modelcard}, Qwen3-VL-235B~
\cite{bai2025qwen3}, Intern-S1 Pro~\cite{zou2026intern}, GLM-4.6V~\cite{hong2025glm}, and Llama-4 Maverick~\cite{adcock2026llama}.
All models are evaluated on identical benchmark instances with the same multimodal inputs and answer-format constraints.


\paragraph{Metrics.}

OmniMatBench includes rubric-based QA and calculation-oriented CAL tasks. 

For QA, each response is evaluated against expert-curated key points and the gold-standard answer. 
We compute Precision, Recall, and F$_1$:
\begin{equation}
\small
\begin{aligned}
P &= \frac{N_{\mathrm{correct}}}{N_{\mathrm{pred}}}, \quad
R = \frac{N_{\mathrm{correct}}}{N_{\mathrm{gold}}}, \quad
F_1 = \frac{2PR}{P+R}.
\end{aligned}
\end{equation}
Here, $N_{\mathrm{correct}}$ is the number of correctly covered key points, $N_{\mathrm{pred}}$ is the number of valid points in the model response, and $N_{\mathrm{gold}}$ is the number of expert-annotated key points. 
Recall measures completeness, while precision measures correctness and penalizes unsupported or irrelevant claims.

For CAL questions, each answer is parsed into structured slots and evaluated using strict exact-slot accuracy:
\begin{equation}
{\small
\mathrm{Acc}_{\mathrm{slot}}
=
\frac{1}{\sum_i |S_i|}
\sum_i \sum_{j \in S_i}
\mathbf{1}[\hat{z}_{ij}=z_{ij}].
}
\end{equation}
Here, $\hat{z}_{ij}=\mathrm{norm}(\hat{y}_{ij})$ and $z_{ij}=\mathrm{norm}(y_{ij})$ denote the normalized (standard) prediction and reference for slot $j$ of item $i$. The overall score is computed as the equally weighted average of CAL exact-slot accuracy and QA Macro-F$_1$:
\begin{equation}
{\small
\mathrm{Score}_{\mathrm{overall}}
=
\frac{1}{2}
\left(
\mathrm{Acc}^{\mathrm{CAL}}_{\mathrm{slot}}
+
\mathrm{F}^{\mathrm{QA}}_{1,\mathrm{macro}}
\right).
}
\end{equation}


\paragraph{Evaluator Validation.}
We mainly use Gemini 2.5 Flash~\cite{comanici2025gemini} as the automated QA evaluator. 
To validate its reliability, we sample 10\% of QA responses from each subfield and compare its scores with expert judgments using the expert-curated key points and gold-standard answers. 
As shown in Table~\ref{tab:spearman}, Spearman's correlations are 0.94 for Recall, 0.91 for Precision, and 0.93 for F1, indicating strong consistency with expert judgments. 
We further conduct MAE-based validation in Appendix Table~\ref{tab:evaluator-validation}, where the F1 MAE is 0.122. 
Together, these results suggest that the automated evaluator is a usable proxy for large-scale QA evaluation.
The overall score is computed as the equally weighted average of CAL exact-slot accuracy and QA Macro-F$_1$.

\begin{table}[!htbp]
\centering
\small
\caption{Spearman's ($\rho$) correlation between human expert and automated evaluator for P, R, and F1.}
\begin{tabularx}{\linewidth}{Xc}
\toprule
\textbf{Metric-Specific Score Correlation} & \textbf{Spearman's $\rho$} \\
\midrule
Recall ($R$) Scores (Completeness) & 0.94 \\
Precision ($P$) Scores (Correctness) & 0.91 \\
F1-Scores & 0.93 \\
\bottomrule
\end{tabularx}
\label{tab:spearman}
\end{table}

\vspace{-0.5em}
\subsection{Results}
\begin{figure}[!htbp]
    \centering
    \includegraphics[width=\columnwidth]{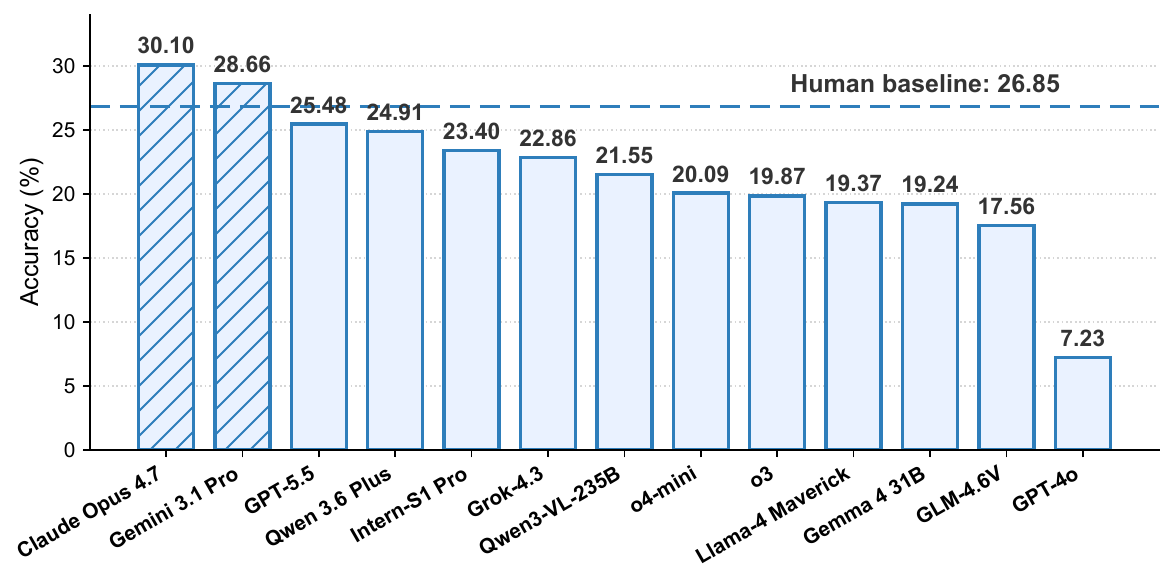}
    \caption{Performance comparison of different models on calculation questions. Slot exact accuracy is reported as the evaluation metric. The dashed line denotes the CAL human reference baseline of 26.85\%.}
    \label{fig:calculation-slot-accuracy}
\end{figure}

\begin{table*}[!htbp]
\centering
\caption{Overall benchmark performance across the 19 OmniMatBench categories. 
The category indices 1--19 correspond to the taxonomy shown in Figure~\ref{fig:benchmark_overview}. 
For each model, the overall score is calculated as the equally weighted average of CAL exact-slot accuracy and QA macro-F1. The human reference baseline is 0.311 overall, computed as the average of QA Macro-F1 (0.353) and CAL strict-slot accuracy (26.85\%). The human reference protocol is described in Appendix~\ref{app:human_baseline}.}
\label{tab:overall-main}

\begingroup
\scriptsize
\setlength{\tabcolsep}{0.8pt}
\setlength{\extrarowheight}{2.4pt}
\renewcommand{\arraystretch}{1.0}
\arrayrulecolor{tabrule}

\resizebox{0.98\textwidth}{!}{%
\begin{tabular}{@{}L{2.45cm}*{19}{C{0.68cm}}C{0.82cm}@{}}
\toprule

\noalign{\vskip -0.55ex}
&
\multicolumn{5}{>{\columncolor{fund}}c}{%
  \color{black}\bfseries
  \rule[1.15ex]{0pt}{3.65ex}%
  \shortstack{Fundamental Materials\\Science}%
}
&
\multicolumn{4}{>{\columncolor{stru}}c}{%
  \color{black}\bfseries
  \rule[1.15ex]{0pt}{3.65ex}%
  \shortstack{Structural and\\Engineering Materials}%
}
&
\multicolumn{6}{>{\columncolor{proc}}c}{%
  \color{black}\bfseries
  \rule[1.15ex]{0pt}{3.65ex}%
  \raisebox{-0.22ex}[0pt][0pt]{\shortstack{Materials Processing\\and Manufacturing}}%
}
&
\multicolumn{4}{>{\columncolor{func}}c}{%
  \color{black}\bfseries
  \rule[1.15ex]{0pt}{3.65ex}%
  \shortstack{Functional and\\Applied Materials}%
}
& \\
\midrule


\rowcolor{gray!8}
\bfseries Model
& \bfseries 1 & \bfseries 2 & \bfseries 3 & \bfseries 4 & \bfseries 5
& \bfseries 6 & \bfseries 7 & \bfseries 8 & \bfseries 9
& \bfseries 10 & \bfseries 11 & \bfseries 12 & \bfseries 13 & \bfseries 14 & \bfseries 15
& \bfseries 16 & \bfseries 17 & \bfseries 18 & \bfseries 19
& \shortstack{\bfseries Avg.\\\bfseries score} \\
\midrule

\textbf{Claude Opus 4.7}
& \scorecell{0.467} & \scorecell{0.510} & \scorecell{0.390} & \scorecell{0.533} & \scorecell{0.423}
& \scorecell{0.414} & \scorecell{0.417} & \scorecell{0.417} & \scorecell{0.320}
& \scorecell{0.557} & \scorecell{0.285} & \scorecell{0.312} & \scorecell{0.278} & \scorecell{0.348} & \scorecell{0.211}
& \scorecell{0.357} & \scorecell{0.333} & \scorecell{0.268} & \scorecell{0.225}
& \avgcell{0.372} \\

\textbf{Gemini 3.1 Pro}
& \scorecell{0.469} & \scorecell{0.454} & \scorecell{0.410} & \scorecell{0.498} & \scorecell{0.418}
& \scorecell{0.426} & \scorecell{0.434} & \scorecell{0.436} & \scorecell{0.299}
& \scorecell{0.527} & \scorecell{0.288} & \scorecell{0.293} & \scorecell{0.216} & \scorecell{0.273} & \scorecell{0.173}
& \scorecell{0.319} & \scorecell{0.212} & \scorecell{0.232} & \scorecell{0.206}
& \avgcell{0.347} \\

\textbf{GPT-5.5}
& \scorecell{0.427} & \scorecell{0.425} & \scorecell{0.382} & \scorecell{0.493} & \scorecell{0.374}
& \scorecell{0.371} & \scorecell{0.394} & \scorecell{0.441} & \scorecell{0.252}
& \scorecell{0.493} & \scorecell{0.251} & \scorecell{0.262} & \scorecell{0.241} & \scorecell{0.259} & \scorecell{0.137}
& \scorecell{0.270} & \scorecell{0.229} & \scorecell{0.282} & \scorecell{0.172}
& \avgcell{0.324} \\

\textbf{Qwen3.6 Plus}
& \scorecell{0.389} & \scorecell{0.406} & \scorecell{0.368} & \scorecell{0.430} & \scorecell{0.369}
& \scorecell{0.342} & \scorecell{0.370} & \scorecell{0.397} & \scorecell{0.260}
& \scorecell{0.472} & \scorecell{0.165} & \scorecell{0.253} & \scorecell{0.202} & \scorecell{0.197} & \scorecell{0.156}
& \scorecell{0.298} & \scorecell{0.185} & \scorecell{0.235} & \scorecell{0.194}
& \avgcell{0.299} \\

\textbf{Grok-4.3}
& \scorecell{0.434} & \scorecell{0.332} & \scorecell{0.365} & \scorecell{0.491} & \scorecell{0.357}
& \scorecell{0.408} & \scorecell{0.383} & \scorecell{0.412} & \scorecell{0.281}
& \scorecell{0.522} & \scorecell{0.214} & \scorecell{0.260} & \scorecell{0.187} & \scorecell{0.131} & \scorecell{0.134}
& \scorecell{0.151} & \scorecell{0.123} & \scorecell{0.163} & \scorecell{0.187}
& \avgcell{0.291} \\

\textbf{o3}
& \scorecell{0.422} & \scorecell{0.366} & \scorecell{0.380} & \scorecell{0.443} & \scorecell{0.351}
& \scorecell{0.377} & \scorecell{0.379} & \scorecell{0.392} & \scorecell{0.285}
& \scorecell{0.425} & \scorecell{0.198} & \scorecell{0.235} & \scorecell{0.190} & \scorecell{0.192} & \scorecell{0.111}
& \scorecell{0.237} & \scorecell{0.127} & \scorecell{0.167} & \scorecell{0.186}
& \avgcell{0.288} \\

\textbf{Gemma-4-31B}
& \scorecell{0.370} & \scorecell{0.350} & \scorecell{0.300} & \scorecell{0.403} & \scorecell{0.310}
& \scorecell{0.287} & \scorecell{0.281} & \scorecell{0.357} & \scorecell{0.192}
& \scorecell{0.364} & \scorecell{0.150} & \scorecell{0.206} & \scorecell{0.139} & \scorecell{0.207} & \scorecell{0.086}
& \scorecell{0.248} & \scorecell{0.186} & \scorecell{0.267} & \scorecell{0.175}
& \avgcell{0.257} \\

\textbf{o4-mini}
& \scorecell{0.397} & \scorecell{0.315} & \scorecell{0.367} & \scorecell{0.419} & \scorecell{0.335}
& \scorecell{0.313} & \scorecell{0.309} & \scorecell{0.377} & \scorecell{0.235}
& \scorecell{0.339} & \scorecell{0.190} & \scorecell{0.189} & \scorecell{0.147} & \scorecell{0.147} & \scorecell{0.064}
& \scorecell{0.202} & \scorecell{0.138} & \scorecell{0.196} & \scorecell{0.144}
& \avgcell{0.254} \\

\textbf{Qwen3-VL-235B}
& \scorecell{0.373} & \scorecell{0.357} & \scorecell{0.300} & \scorecell{0.443} & \scorecell{0.301}
& \scorecell{0.300} & \scorecell{0.300} & \scorecell{0.361} & \scorecell{0.228}
& \scorecell{0.416} & \scorecell{0.172} & \scorecell{0.193} & \scorecell{0.154} & \scorecell{0.157} & \scorecell{0.075}
& \scorecell{0.149} & \scorecell{0.125} & \scorecell{0.148} & \scorecell{0.140}
& \avgcell{0.247} \\

\textbf{Intern-S1 Pro}
& \scorecell{0.362} & \scorecell{0.274} & \scorecell{0.326} & \scorecell{0.449} & \scorecell{0.294}
& \scorecell{0.278} & \scorecell{0.284} & \scorecell{0.383} & \scorecell{0.212}
& \scorecell{0.384} & \scorecell{0.170} & \scorecell{0.210} & \scorecell{0.187} & \scorecell{0.157} & \scorecell{0.073}
& \scorecell{0.163} & \scorecell{0.150} & \scorecell{0.115} & \scorecell{0.156}
& \avgcell{0.243} \\

\textbf{GLM-4.6V}
& \scorecell{0.319} & \scorecell{0.263} & \scorecell{0.282} & \scorecell{0.362} & \scorecell{0.283}
& \scorecell{0.261} & \scorecell{0.241} & \scorecell{0.304} & \scorecell{0.228}
& \scorecell{0.360} & \scorecell{0.146} & \scorecell{0.143} & \scorecell{0.144} & \scorecell{0.142} & \scorecell{0.043}
& \scorecell{0.137} & \scorecell{0.113} & \scorecell{0.135} & \scorecell{0.128}
& \avgcell{0.212} \\

\textbf{Llama-4 Maverick}
& \scorecell{0.336} & \scorecell{0.236} & \scorecell{0.263} & \scorecell{0.388} & \scorecell{0.276}
& \scorecell{0.282} & \scorecell{0.268} & \scorecell{0.292} & \scorecell{0.151}
& \scorecell{0.335} & \scorecell{0.148} & \scorecell{0.111} & \scorecell{0.127} & \scorecell{0.131} & \scorecell{0.035}
& \scorecell{0.106} & \scorecell{0.105} & \scorecell{0.129} & \scorecell{0.080}
& \avgcell{0.200} \\

\textbf{GPT-4o}
& \scorecell{0.265} & \scorecell{0.152} & \scorecell{0.187} & \scorecell{0.293} & \scorecell{0.197}
& \scorecell{0.211} & \scorecell{0.226} & \scorecell{0.217} & \scorecell{0.155}
& \scorecell{0.315} & \scorecell{0.095} & \scorecell{0.103} & \scorecell{0.049} & \scorecell{0.058} & \scorecell{0.021}
& \scorecell{0.065} & \scorecell{0.032} & \scorecell{0.044} & \scorecell{0.083}
& \avgcell{0.146} \\
\bottomrule
\end{tabular}%
}

\vspace{0.8mm}
\heatmaplegend

\arrayrulecolor{black}
\endgroup
\end{table*}
We conducted a comprehensive evaluation of 13 leading MLLMs on the \textit{OmniMatBench} benchmark. 
Table~\ref{tab:overall-main} reports the category-level overall scores across the 19 materials-science subfields. 
Table~\ref{tab:qa-4class} provides detailed QA results in terms of Precision, Recall, and F$_1$ across the four major OmniMat groups. 
Figure~\ref{fig:calculation-slot-accuracy} summarizes model performance on CAL questions using strict slot exact accuracy. 
Overall, the results reveal several key findings.

\paragraph{Observation 1: Overall performance and model hierarchy.}
As shown in Table~\ref{tab:overall-main}, Claude Opus 4.7 achieves the best overall score of 0.372, followed by Gemini 3.1 Pro with 0.347 and GPT-5.5 with 0.324. 
Most other models obtain overall scores below 0.30, indicating that broad materials-science reasoning remains challenging even for frontier MLLMs. 
Closed-source models generally outperform open-source models. 
The best open-source model, Gemma-4-31B, achieves an overall score of 0.257, still trailing Claude Opus 4.7 by 0.115. 
We further evaluate small vision-language models on OmniMatBench, as shown in Table~\ref{tab:small_model_cal_ablation} in the appendix. 
Notably, Qwen3-VL-8B-Thinking improves over Qwen3-VL-8B-Instruct by 5.55 points, suggesting that explicit reasoning-oriented inference is beneficial for materials-science tasks. 
However, all small models still obtain accuracy below 0.20, confirming the difficulty of OmniMatBench for compact models.
This suggests that OmniMatBench remains unsaturated and can distinguish models across families.

\vspace{-1em}

\paragraph{Observation 2: Performance varies substantially across materials-science subfields.}
The category-level results in Table~\ref{tab:overall-main} show that models perform better on some fundamental or well-structured topics, such as metallic materials, materials chemistry, and welding technology and engineering. 
In contrast, application-oriented and engineering-intensive areas, such as metallurgical engineering, electronic information materials, nanomaterials and technology, and composite materials molding, consistently receive lower scores. 
This pattern suggests that models handle common materials-science concepts better, but struggle with specialized materials-engineering scenarios requiring procedural knowledge, materials-specific formulas, and application-oriented judgment.

\begin{table*}[!htbp]
\centering
\caption{Comparison of different MLLMs on 4 coarse-grained materials-science benchmark categories. \textbf{P}, \textbf{R}, and \textbf{F1} represent Precision, Recall, and Macro F1-score, respectively. \textbf{FMS}: Fundamental Materials Science, \textbf{SEM}: Structural and Engineering Materials, \textbf{MPM}: Materials Processing and Manufacturing, \textbf{FAM}: Functional and Applied Materials. The best results are in \textbf{bold} and the second best are \underline{underlined}. The overall human F1-score baseline is 0.353.}
\label{tab:qa-4class}
\resizebox{\textwidth}{!}{%
\begin{tabular}{lccc|ccc|ccc|ccc|ccc}
\toprule
\textbf{Model} 
& \multicolumn{3}{c}{\textbf{FMS}} 
& \multicolumn{3}{c}{\textbf{SEM}} 
& \multicolumn{3}{c}{\textbf{MPM}} 
& \multicolumn{3}{c}{\textbf{FAM}} 
& \multicolumn{3}{c}{\textbf{Overall}} \\
\cmidrule(lr){2-4}
\cmidrule(lr){5-7}
\cmidrule(lr){8-10}
\cmidrule(lr){11-13}
\cmidrule(lr){14-16}
& \textbf{P} & \textbf{R} & \textbf{F1}
& \textbf{P} & \textbf{R} & \textbf{F1}
& \textbf{P} & \textbf{R} & \textbf{F1}
& \textbf{P} & \textbf{R} & \textbf{F1}
& \textbf{P} & \textbf{R} & \textbf{F1} \\
\midrule
\multicolumn{16}{c}{\textit{Closed-source Models}} \\
\midrule
Claude Opus 4.7 
& 0.517 & \textbf{0.727} & \textbf{0.540}
& 0.405 & \textbf{0.675} & \underline{0.444}
& 0.477 & \textbf{0.461} & \textbf{0.384}
& \underline{0.520} & \textbf{0.461} & \textbf{0.407}
& 0.481 & \textbf{0.576} & \textbf{0.443} \\
Gemini 3.1 Pro 
& 0.514 & \underline{0.685} & \underline{0.521}
& 0.421 & 0.616 & 0.434
& \underline{0.479} & 0.393 & \underline{0.346}
& \underline{0.520} & 0.362 & 0.327
& \underline{0.485} & 0.510 & \underline{0.406} \\
GPT-5.5 
& 0.465 & 0.655 & 0.482
& 0.393 & 0.585 & 0.404
& 0.413 & \underline{0.424} & 0.333
& 0.466 & \underline{0.408} & \underline{0.361}
& 0.434 & \underline{0.515} & 0.393 \\
o3 
& 0.503 & 0.632 & 0.503
& 0.433 & \underline{0.623} & \textbf{0.450}
& 0.407 & 0.368 & 0.303
& 0.393 & 0.292 & 0.255
& 0.435 & 0.475 & 0.376 \\
Grok-4.3 
& \textbf{0.599} & 0.551 & 0.489
& \textbf{0.480} & 0.552 & 0.440
& \textbf{0.495} & 0.312 & 0.294
& \textbf{0.536} & 0.200 & 0.189
& \textbf{0.528} & 0.402 & 0.354 \\
Qwen3.6 Plus 
& 0.429 & 0.671 & 0.455
& 0.349 & 0.618 & 0.372
& 0.396 & 0.351 & 0.278
& 0.461 & 0.369 & 0.302
& 0.409 & 0.495 & 0.350 \\
o4-mini 
& 0.521 & 0.552 & 0.458
& 0.386 & 0.508 & 0.353
& 0.361 & 0.262 & 0.218
& 0.411 & 0.206 & 0.207
& 0.419 & 0.378 & 0.307 \\
GPT-4o 
& 0.500 & 0.370 & 0.339
& 0.423 & 0.381 & 0.326
& 0.346 & 0.159 & 0.152
& 0.267 & 0.052 & 0.062
& 0.386 & 0.239 & 0.219 \\
\midrule
\multicolumn{16}{c}{\textit{Open-source Models}} \\
\midrule
Gemma-4-31B 
& 0.424 & 0.606 & 0.420
& 0.316 & 0.569 & 0.330
& 0.335 & 0.310 & 0.232
& 0.430 & 0.368 & 0.324
& 0.375 & 0.455 & 0.321 \\
Qwen3-VL-235B 
& 0.521 & 0.473 & 0.426
& 0.414 & 0.424 & 0.336
& 0.370 & 0.223 & 0.209
& 0.424 & 0.123 & 0.141
& 0.431 & 0.310 & 0.278 \\
Intern-S1 Pro 
& 0.516 & 0.440 & 0.388
& 0.394 & 0.406 & 0.318
& 0.366 & 0.205 & 0.192
& 0.424 & 0.119 & 0.111
& 0.424 & 0.291 & 0.253 \\
GLM-4.6V 
& 0.438 & 0.456 & 0.365
& 0.352 & 0.446 & 0.309
& 0.332 & 0.222 & 0.185
& 0.375 & 0.148 & 0.140
& 0.373 & 0.315 & 0.249 \\
Llama-4 Maverick 
& 0.537 & 0.371 & 0.338
& 0.394 & 0.352 & 0.287
& 0.387 & 0.140 & 0.135
& 0.368 & 0.056 & 0.067
& 0.424 & 0.228 & 0.206 \\

\bottomrule
\end{tabular}%
}
\end{table*}

\paragraph{Observation 3: Broad Coverage but Limited Precision in Open-Ended QA.}
Table~\ref{tab:qa-4class} shows a common pattern: many models achieve higher Recall than Precision in open-ended QA. 
For instance, Claude Opus 4.7 obtains an overall Recall of 0.576 but a Precision of 0.481, suggesting that its responses cover many materials-science key points while still including unsupported or less relevant content. 
This trend indicates that MLLMs can often generate broad materials-related answers, but they still lack precise control over which scientific points are truly relevant to the question. 
Thus, for some materials QA tasks, models still struggle to identify the most relevant scientific key points.


\paragraph{Observation 4: CAL questions expose a stronger execution bottleneck.}
As shown in Figure~\ref{fig:calculation-slot-accuracy}, the best CAL slot exact accuracy is only 30.10\%, achieved by Claude Opus 4.7. 
Gemini 3.1 Pro follows with 28.66\%, while GPT-5.5 reaches 25.48\%. 
Only the top two models exceed the human baseline of 26.85\%, and most models remain below 25\%. 
This indicates that calculation-oriented materials reasoning is substantially harder than producing plausible explanations~\cite{liu2026polyreal}. 
Therefore, current MLLMs still exhibit a clear gap between scientific fluency and reliable scientific execution.

\vspace{-0.2em}
\section{Analysis}
We further analyze model failures from three perspectives: representative case studies, formula/retrieval assistance, and code execution.

\subsection{Case Studies: Failures in Materials-Specific Reasoning}

\begin{figure*}[!htbp]
    \centering
    \includegraphics[width=\textwidth]{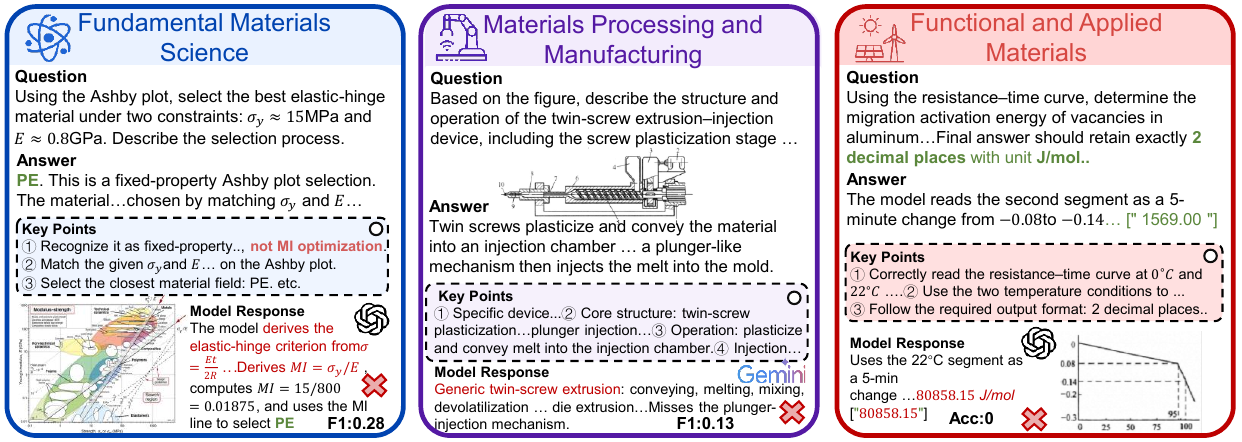}
    \caption{Representative error cases in OmniMatBench. 
The examples show failures across the knowledge-to-execution process, including wrong material-selection criteria, missing device mechanisms, graph-reading errors, and incorrect numerical execution.
}
    \label{fig:case}
\end{figure*}

Figure~\ref{fig:case} presents three representative failure cases from different OmniMat domains. 
These cases show that model errors are not limited to missing final answers, but often involve incorrect reasoning paths, incomplete materials-process understanding, and failures in scientific execution.

\paragraph{Case 1: Correct Answer with Incorrect Materials Reasoning.}
The first case comes from Fundamental Materials Science and asks the model to select the best elastic-hinge material from an Ashby plot under given constraints on yield strength and elastic modulus. 
The correct solution is a fixed-property Ashby plot selection: the model should match the given $\sigma_y$ and $E$ values to the closest material region and select PE. 
However, the model over-applies a common material-index formulation for elastic-hinge optimization and derives an unnecessary criterion before reaching the final answer. 
This reveals a tendency of MLLMs to follow familiar materials-design heuristics rather than adapt their reasoning to the task.
Importantly, the expert-curated key points explicitly indicate that this is not an MI optimization problem, allowing OmniMatBench to penalize superficially correct answers with incorrect reasoning.

\paragraph{Case 2: Generic Materials-Processing Knowledge Instead of Process-Specific Understanding.}
The second case is from Materials Processing and Manufacturing and requires understanding the structure and operation of a twin-screw extrusion--injection device from the figure. 
A correct answer should describe the twin screws feeding the melt into an injection chamber, followed by injection into the mold. 
The model instead gives a generic description of twin-screw extrusion and misses the injection mechanism.
This suggests that models may possess broad materials-processing vocabulary but lack precise understanding of specialized processing equipment and operation sequences. 
Such failures highlight the importance of covering subfields such as materials processing and manufacturing in OmniMatBench.

\begin{table}[!htbp]
\centering
\caption{Macro-average performance on 458 questions sampled from OmniMat under retrieval settings. Parentheses denote absolute improvements over Vanilla.}
\label{tab:matrag_macro_results}
\tiny
\setlength{\tabcolsep}{3.8pt}
\resizebox{0.95\columnwidth}{!}{
\begin{tabular}{llc}
\toprule
\textbf{Model} & \textbf{Setting} & \textbf{Macro Avg.} \\
\midrule

\multirow{3}{*}{Gemini-3.1-Pro}
& Vanilla & 25.47 \\
& + Oracle Context & \textbf{35.62} {\tiny (+10.15)} \\
& + Distractor Retrieval & 33.53 {\tiny (+8.06)} \\
\midrule

\multirow{3}{*}{GPT-5.5}
& Vanilla & 23.37 \\
& + Oracle Context & \textbf{27.98} {\tiny (+4.61)} \\
& + Distractor Retrieval & 26.01 {\tiny (+2.64)} \\
\midrule

\multirow{3}{*}{o3}
& Vanilla & 19.73 \\
& + Oracle Context & \textbf{25.92} {\tiny (+6.19)} \\
& + Distractor Retrieval & 23.65 {\tiny (+3.92)} \\
\midrule

\multirow{3}{*}{Intern-S1-Pro}
& Vanilla & 25.67 \\
& + Oracle Context & \textbf{30.18} {\tiny (+4.51)} \\
& + Distractor Retrieval & 27.83 {\tiny (+2.16)} \\
\bottomrule
\end{tabular}
}
\end{table}
\paragraph{Case 3: Visual-Parameter Misinterpretation in Applied Materials Calculation.}
The third case is from Functional and Applied Materials and requires computing the vacancy migration activation energy in aluminum from a resistance--time curve. 
Solving the problem requires correct interpretation of the plotted segments, temperature conditions, formula selection, and parameter substitution. 
The model misreads the relevant curve segment and applies an incorrect calculation path, producing a value with the wrong magnitude. 
Because OmniMatBench specifies the required unit and numerical precision for each CAL item, this case exposes that the error is not merely a formatting mismatch, but a failure to connect materials-specific visual information with the appropriate physical calculation.

Together, these cases show that model failures are not limited to missing knowledge. 
Models may obtain the right answer for the wrong reason, replace specialized materials-processing knowledge with generic descriptions, or fail during visual interpretation, formula selection, and parameter extraction. 
These failures justify the design of OmniMatBench, which combines expert-curated QA key points with structured CAL answer slots to evaluate materials understanding and scientific execution.


\subsection{Formula Assistance Probes Materials-Knowledge Application}

We further use formula-level assistance as method-level materials knowledge to examine whether models can apply the correct solution plan in CAL reasoning. 
Table~\ref{tab:matrag_macro_results} reports results on a sampled subset of 458 questions, comparing Vanilla inference with Oracle Context and Distractor Retrieval. 
In the Vanilla setting, models solve the problems without additional formula guidance. 
Oracle Context provides the ground-truth formula plan, testing whether models can apply a correct high-level materials solution path when it is given. 
Distractor Retrieval provides the correct formula plan together with similar distractor plans, testing whether models can identify the applicable materials-specific solution plan from plausible alternatives.

Formula assistance improves all evaluated models under Oracle Context: Gemini 3.1 Pro rises from 25.47 to 35.62, o3 from 19.73 to 25.92, and Intern-S1 Pro from 25.67 to 30.18. Smaller gains under Distractor Retrieval indicate that models still struggle to choose the correct materials-specific solution plan among plausible alternatives.

Oracle Context does not saturate performance, with the best score remaining 35.62. Formula access alone is insufficient: models must judge applicability, ground variables from multimodal inputs, and adapt the solution plan to each scenario.



Therefore, the remaining gap suggests that the bottleneck is not access to formulas, but the ability to understand and apply method-level materials knowledge under problem-specific constraints.

\subsection{Code Execution Reveals Uneven Materials-Knowledge Application}


We further test whether external code execution improves materials CAL reasoning using 150 questions sampled from Materials Physics, Materials Chemistry, and Functional Materials. Table~\ref{tab:code_macro_results} compares Vanilla inference, Code Execution with externally executed model-generated Python code, and Code w/ Gold Target, which additionally provides the correct formula plan before code generation.





\begin{table}[!htbp]
\centering
\caption{Macro-average performance on 150 sampled OmniMat subfield questions under code assistance. Parentheses denote changes relative to Vanilla.}

\label{tab:code_macro_results}
\tiny
\setlength{\tabcolsep}{3.8pt}
\resizebox{0.95\columnwidth}{!}{
\begin{tabular}{llc}
\toprule
\textbf{Model} & \textbf{Setting} & \textbf{Macro Avg.} \\
\midrule

\multirow{3}{*}{GPT-5.5}
& Vanilla & \textbf{27.65} \\
& + Code Execution & 19.06 {\tiny (-8.60)} \\
& + Code w/ Gold Target & 21.59 {\tiny (-6.07)} \\
\midrule

\multirow{3}{*}{o3}
& Vanilla & \textbf{23.67} \\
& + Code Execution & 18.42 {\tiny (-5.25)} \\
& + Code w/ Gold Target & 22.25 {\tiny (-1.42)} \\
\midrule

\multirow{3}{*}{Intern-S1 Pro}
& Vanilla & 27.69 \\
& + Code Execution & 25.38 {\tiny (-2.31)} \\
& + Code w/ Gold Target & \textbf{27.82} {\tiny (+0.13)} \\
\bottomrule
\end{tabular}
}
\end{table}

Code-augmented inference does not consistently improve performance. 
GPT-5.5 and o3 both drop under Code Execution, from 27.65 to 19.06 and from 23.67 to 18.42, respectively. 
Providing the gold target partially recovers performance, but still remains below Vanilla for both models. 
Intern-S1 Pro is more stable, with Code w/ Gold Target slightly improving its score from 27.69 to 27.82.

These results suggest that the main bottleneck is not simply arithmetic computation. 
In many cases, the generated code can be executed, but models still fail to translate the materials problem into the correct computational procedure. 
Errors often arise from misunderstanding the materials context, selecting wrong parameters, mapping quantities incorrectly from the question or figure, or substituting values into the wrong formula. 
Thus, code execution can only help when the model first applies the relevant materials knowledge correctly.

\begin{figure}[!htbp]
    \centering
    \includegraphics[width=0.97\linewidth]{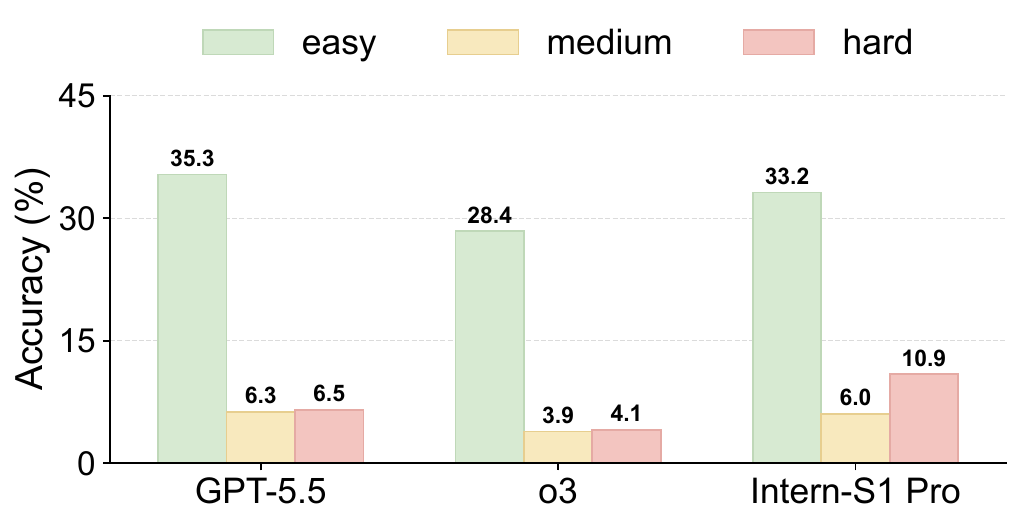}
    \caption{CAL Accuracy Across Difficulty Levels}
    \label{fig:calculation-slot-accuracy-partial-difficulty}
\end{figure}
The results for Intern-S1 Pro suggest that scientific-domain specialization can help in specific materials subfields. On this subset, it slightly outperforms GPT-5.5 in the Vanilla setting, remains more stable under code-assisted settings, and achieves the highest accuracy on hard CAL questions in Figure~\ref{fig:calculation-slot-accuracy-partial-difficulty}. This indicates that specialization may help models use code more effectively when the correct solution plan is provided. However, the gains remain limited: reliable scientific execution requires more than runnable code; models must understand the materials scenario, extract variables, and apply domain-specific materials knowledge.


\section{Conclusion}

We presented \textsc{OmniMatBench}, a human-calibrated multimodal reasoning benchmark for materials science. It contains 3,171 expert-curated problems across 19 materials-science subfields, spanning fundamental materials knowledge, structural and engineering materials, materials processing and manufacturing, and functional and applied materials. \textsc{OmniMatBench} evaluates MLLMs through rubric-based QA and calculation-oriented CAL tasks, covering materials understanding and scientific execution. We systematically evaluated 13 open-source and closed-source MLLMs, showing that the benchmark remains far from saturated: models perform better on common materials concepts but still struggle with specialized engineering scenarios, process-specific mechanisms, formula-plan selection, visual-parameter grounding, and structured scientific computation. These findings highlight the gap between scientific fluency and reliable materials-science reasoning and execution.







\section*{Limitations}

Although \textsc{OmniMatBench} covers 3,171 expert-curated problems across 19 materials-science subfields, it remains a bounded sample of materials-science reasoning. We screened and balanced problems as much as possible while preserving evaluation quality, but variations in available domain knowledge still lead to some imbalance in question numbers and topic coverage. Some underrepresented subfields, such as nanomaterials and technology, and gem materials and gemology, are often overlooked by existing materials benchmarks but remain important for materials research and engineering applications. We therefore include these areas while maintaining the best possible balance under available high-quality sources.

\textsc{OmniMatBench} focuses on expert-curated QA and calculation-oriented CAL tasks with reference solutions, rubrics, formula traces, and structured answer slots. QA scoring key points are cross-validated by materials-domain experts to improve consistency and reduce ambiguity. However, these expert-defined rubrics may still reflect subjective judgments and may underweight valid reasoning elements outside the annotated criteria. Future work can extend \textsc{OmniMatBench} to richer scientific modalities and more flexible evaluation methods.




\section*{Ethical considerations}

OmniMatBench was constructed with the participation of 73 materials-domain annotators and validators, including domain experts and trained graduate-level contributors. 
All contributors were informed of the purpose of the project and the intended use of their annotations for research and benchmark construction. 
They were compensated according to local institutional standards and the amount of work completed. 
The annotation process involved problem extraction, domain adaptation, proof-file construction, key-point annotation, and expert validation. 
No personal or sensitive information about contributors is released, and the benchmark does not contain private user data. 
The source materials were used only to construct expert-curated problem--solution pairs and evaluation rubrics, and all released items are reviewed to ensure scientific correctness, answer consistency, and appropriate use for academic evaluation. 
We further apply both LLM-based verification and human expert validation to reduce noise, redundancy, and potential annotation errors.

\section*{Acknowledgements}

This work is supported by New Generation Artificial Intelligence-National Science and Technology Major Project(2025ZD0121802) and Intern-Discovery. This work was done during the authors' internships at Shanghai Artificial Intelligence Laboratory.




\bibliography{custom}

\appendix
\clearpage
\section{AI Assistance Statement}

The authors used AI-based tools only for auxiliary writing support, including grammar checking, wording refinement, and improving readability. These tools were not used to determine the research questions, construct the benchmark, design the evaluation protocol, implement the experiments, analyze the results, or formulate the scientific claims. All AI-assisted edits were manually reviewed, verified, and revised by the authors, who take full responsibility for the final content of the paper.

\section{Data statistics}
Table~\ref{app:data-statistics} reports the category-level distribution of QA and CAL questions in OmniMatBench. 
The 19 categories cover four major groups of materials science, including fundamental materials science, structural and engineering materials, materials processing and manufacturing, and functional and applied materials. 
For each category, we report the number of QA questions, CAL questions, and the total number of benchmark instances. 
Overall, OmniMatBench contains 1,261 QA questions and 1,910 CAL questions, resulting in 3,171 questions in total.

\begin{table}[!htbp]
\centering
\caption{Data statistics across the 19 OmniMatBench categories. Each category contains both QA and CAL questions, and the benchmark includes 3,171 questions in total.}
\label{app:data-statistics}

\begingroup
\small
\setlength{\tabcolsep}{5.5pt}
\renewcommand{\arraystretch}{1.08}

\resizebox{\linewidth}{!}{%
\begin{tabular}{@{}clrrr@{}}
\toprule
\textbf{ID} & \textbf{Category} & \textbf{QA} & \textbf{CAL} & \textbf{Total} \\
\midrule
1  & Materials Science and Engineering        & 85  & 78  & 163 \\
2  & Soft Matter Science and Engineering      & 70  & 64  & 134 \\
3  & Materials Physics                        & 75  & 91  & 166 \\
4  & Materials Chemistry                      & 80  & 96  & 176 \\
5  & Materials Design Science and Engineering & 78  & 87  & 165 \\
6 & Polymer Materials                        & 62  & 99  & 161 \\
7 & Composite Materials                      & 80  & 88  & 168 \\
8 & Powder Materials                         & 81  & 101 & 182 \\
9 & Gem Materials and Gemology               & 87  & 75  & 162 \\
10  & Metallurgical Engineering                & 96  & 86  & 182 \\
11  & Welding Technology and Engineering       & 84  & 87  & 171 \\
12  & Intelligent Materials and Structures     & 36  & 168 & 204 \\
13  & New Energy Materials and Devices         & 16  & 166 & 182 \\
14 & Metallic Materials                       & 86  & 95  & 181 \\
15 & Inorganic Non-metallic Materials         & 44  & 87  & 131 \\
16 & Functional Materials                     & 35  & 131 & 166 \\
17 & Electronic Information Materials         & 91  & 140 & 231 \\
18 & Nanomaterials and Technology             & 11  & 87  & 98  \\
19 & Composite Materials Molding              & 64  & 84  & 148 \\
\midrule
\multicolumn{2}{@{}l}{\textbf{Total}} & \textbf{1261} & \textbf{1910} & \textbf{3171} \\
\bottomrule
\end{tabular}%
}

\endgroup
\end{table}
\label{sec:appendix_prompt}

\section{Additional Findings on CAL}
\label{app:cal_additional_findings}

Table~\ref{tab:cal-strict-slot-main-highlight} further reveals two category-level patterns that are less visible from the overall averages.

\paragraph{Performance is highly category-dependent.}
CAL performance is not uniformly distributed across the 19 categories.
Models generally obtain higher scores on chemistry- and processing-centered categories, such as Materials Chemistry, Welding Technology and Engineering, Metallic Materials, and Powder Materials.
For example, several strong models achieve their peak or near-peak scores on Powder Materials, including GPT-5.5, Gemini 3.1 Pro, and Qwen3.6 Plus.
In contrast, Metallurgical Engineering, Functional Materials, Electronic Information Materials, and Composite Materials Forming Engineering remain consistently difficult.
This suggests that CAL is especially challenging when the required slots involve implicit process--property relationships, functional mechanisms, or application-specific roles, rather than explicit material names or processing descriptors.

\paragraph{Similar average scores can hide different model behaviors.}
The category-wise results also show that models with similar average CAL scores may rely on different strengths.
For instance, Claude Opus 4.7 and Gemini 3.1 Pro obtain close overall averages, but their category profiles differ: Gemini 3.1 Pro is stronger in several fundamental-materials categories, whereas Claude Opus 4.7 shows advantages in multiple structural, processing, and applied-materials categories.
Likewise, Intern-S1 Pro is the best open-source model overall, but its competitiveness is concentrated in selected categories rather than being uniform across all subdomains.
These differences indicate that aggregate CAL scores should be interpreted together with category-level results, since the same average performance may correspond to distinct forms of materials-domain competence.
\begin{table*}[!htbp]
\centering
\caption{CAL strict-slot performance across the 19 OmniMatBench subfields. The top, second-best, and lowest scores in each column are highlighted in red, yellow, and blue, respectively.}
\label{tab:cal-strict-slot-main-highlight}

\begingroup
\scriptsize
\definecolor{bestred}{HTML}{F4CCCC}
\definecolor{secondyellow}{HTML}{FFF2CC}
\definecolor{worstblue}{HTML}{D9EAF7}
\setlength{\tabcolsep}{1.45pt}
\setlength{\extrarowheight}{2.1pt}
\setlength{\arrayrulewidth}{0.35pt}
\renewcommand{\arraystretch}{1.0}

\resizebox{\textwidth}{!}{%
\begin{tabular}{@{}L{2.55cm}|*{5}{C{0.68cm}}|*{4}{C{0.68cm}}|*{6}{C{0.68cm}}|*{4}{C{0.68cm}}|C{0.82cm}@{}}
\toprule
\multicolumn{1}{c|}{\bfseries Models} & 
\multicolumn{5}{C{3.4cm}|}{\bfseries Fundamental Materials Science} & 
\multicolumn{4}{C{2.7cm}|}{\bfseries Structural and Engineering Materials} & 
\multicolumn{6}{C{4.0cm}|}{\bfseries Materials Processing and Manufacturing} & 
\multicolumn{4}{C{2.7cm}|}{\bfseries Functional and Applied Materials} & 
\bfseries Avg. \\
\midrule
\bfseries Model & \bfseries 1 & \bfseries 2 & \bfseries 3 & \bfseries 4 & \bfseries 5 & \bfseries 6 & \bfseries 7 & \bfseries 8 & \bfseries 9 & \bfseries 10 & \bfseries 11 & \bfseries 12 & \bfseries 13 & \bfseries 14 & \bfseries 15 & \bfseries 16 & \bfseries 17 & \bfseries 18 & \bfseries 19 & \bfseries Avg. \\
\midrule
\multicolumn{21}{c}{\itshape Closed-source MLLMs} \\
\midrule
\textbf{Claude Opus 4.7} & \cellcolor{secondyellow}0.348 & \cellcolor{secondyellow}0.138 & 0.263 & \cellcolor{bestred}0.478 & 0.279 & \cellcolor{secondyellow}0.149 & \cellcolor{bestred}0.497 & \cellcolor{bestred}0.217 & \cellcolor{bestred}0.271 & 0.413 & \cellcolor{bestred}0.424 & \cellcolor{secondyellow}0.342 & \cellcolor{secondyellow}0.407 & 0.373 & 0.211 & \cellcolor{secondyellow}0.142 & \cellcolor{bestred}0.227 & 0.206 & 0.151 & \cellcolor{bestred}0.291 \\
\textbf{Gemini 3.1 Pro} & \cellcolor{bestred}0.376 & \cellcolor{bestred}0.153 & \cellcolor{bestred}0.333 & 0.444 & \cellcolor{bestred}0.327 & \cellcolor{bestred}0.162 & 0.441 & 0.140 & 0.195 & \cellcolor{secondyellow}0.418 & \cellcolor{secondyellow}0.372 & \cellcolor{bestred}0.347 & \cellcolor{bestred}0.433 & \cellcolor{secondyellow}0.470 & 0.204 & 0.113 & 0.129 & \cellcolor{secondyellow}0.216 & \cellcolor{secondyellow}0.174 & \cellcolor{secondyellow}0.287 \\
\textbf{GPT-5.5} & 0.308 & 0.095 & \cellcolor{secondyellow}0.318 & \cellcolor{secondyellow}0.449 & \cellcolor{secondyellow}0.322 & 0.110 & \cellcolor{secondyellow}0.472 & 0.130 & 0.175 & \cellcolor{bestred}0.428 & 0.273 & 0.276 & 0.368 & \cellcolor{bestred}0.503 & 0.150 & 0.083 & 0.084 & 0.127 & 0.168 & 0.255 \\
\textbf{Qwen3.6 Plus} & 0.175 & 0.124 & 0.288 & 0.404 & 0.293 & 0.143 & 0.429 & 0.181 & 0.164 & 0.365 & 0.299 & 0.286 & 0.368 & 0.456 & 0.156 & 0.108 & 0.114 & 0.196 & \cellcolor{bestred}0.179 & 0.249 \\
\textbf{Grok-4.3} & 0.235 & 0.130 & 0.242 & 0.404 & 0.226 & 0.084 & 0.410 & 0.117 & 0.153 & 0.361 & 0.277 & 0.276 & 0.303 & 0.413 & 0.218 & 0.078 & 0.116 & 0.162 & 0.138 & 0.229 \\
\textbf{o3} & 0.213 & 0.089 & 0.268 & 0.365 & 0.192 & 0.065 & 0.292 & 0.053 & 0.167 & 0.351 & 0.234 & 0.236 & 0.268 & 0.403 & 0.170 & 0.073 & 0.084 & 0.127 & 0.125 & 0.199 \\
\textbf{o4-mini} & 0.222 & 0.098 & 0.227 & 0.343 & 0.216 & 0.032 & 0.217 & 0.119 & 0.158 & 0.356 & 0.238 & 0.226 & 0.273 & 0.376 & 0.184 & 0.094 & 0.127 & 0.191 & 0.115 & 0.201 \\
\textbf{GPT-4o} & \cellcolor{worstblue}0.122 & \cellcolor{worstblue}0.051 & \cellcolor{worstblue}0.101 & \cellcolor{worstblue}0.152 & \cellcolor{worstblue}0.043 & \cellcolor{worstblue}0.000 & \cellcolor{worstblue}0.139 & \cellcolor{worstblue}0.013 & \cellcolor{worstblue}0.045 & \cellcolor{worstblue}0.133 & \cellcolor{worstblue}0.064 & \cellcolor{worstblue}0.035 & \cellcolor{worstblue}0.064 & \cellcolor{worstblue}0.106 & \cellcolor{worstblue}0.107 & \cellcolor{worstblue}0.045 & \cellcolor{worstblue}0.043 & \cellcolor{worstblue}0.071 & \cellcolor{worstblue}0.040 & \cellcolor{worstblue}0.072 \\
\midrule
\multicolumn{21}{c}{\itshape Open-source MLLMs} \\
\midrule
\textbf{Intern-S1 Pro} & 0.243 & \cellcolor{secondyellow}0.138 & 0.280 & 0.404 & 0.226 & 0.077 & 0.323 & \cellcolor{secondyellow}0.185 & \cellcolor{secondyellow}0.243 & 0.356 & 0.203 & 0.201 & 0.229 & 0.430 & 0.156 & \cellcolor{bestred}0.144 & \cellcolor{secondyellow}0.189 & \cellcolor{bestred}0.230 & 0.151 & 0.232 \\
\textbf{Qwen3-VL-235B} & 0.231 & 0.118 & 0.237 & 0.393 & 0.202 & 0.065 & 0.168 & 0.167 & 0.198 & 0.317 & 0.260 & 0.201 & 0.225 & 0.373 & \cellcolor{secondyellow}0.231 & 0.105 & 0.157 & 0.176 & 0.102 & 0.207 \\
\textbf{Gemma-4-31B} & 0.175 & 0.115 & 0.212 & 0.331 & 0.216 & 0.065 & 0.292 & 0.108 & 0.158 & 0.356 & 0.251 & 0.191 & 0.221 & 0.373 & 0.136 & 0.070 & 0.099 & 0.118 & 0.166 & 0.192 \\
\textbf{Llama-4 Maverick} & 0.243 & 0.120 & 0.227 & 0.326 & 0.187 & 0.019 & 0.230 & 0.152 & 0.201 & 0.293 & 0.273 & 0.221 & 0.162 & 0.313 & 0.136 & 0.102 & 0.160 & \cellcolor{bestred}0.230 & 0.078 & 0.193 \\
\textbf{GLM-4.6V} & 0.192 & 0.071 & 0.242 & 0.309 & 0.174 & 0.026 & 0.238 & 0.153 & 0.166 & 0.252 & 0.212 & 0.171 & 0.149 & 0.250 & \cellcolor{bestred}0.238 & 0.102 & 0.128 & 0.160 & 0.074 & 0.174 \\
\bottomrule
\end{tabular}%
}


\endgroup
\end{table*}

\section{Detailed Benchmark Analysis}
\label{sec:formula-tool}

OmniMatBench evaluates not only whether a model can answer a materials question, but also whether it can complete a case-based reasoning work-up: identify the relevant operation, apply it to the current problem, and output the result in the required answer slots. We therefore evaluate several inference-time assistance settings while keeping the benchmark questions, multimodal inputs, answer-slot constraints, and evaluation metrics unchanged.

\paragraph{Zero-shot inference}
In the zero-shot setting, the model receives the problem statement, optional image or CSV input, and the final-answer schema. The model is instructed to place its reasoning in a \texttt{<think>} block and to output only the final filled answer structure inside \texttt{<answer>...</answer>}. If a \texttt{final\_answer\_format} is provided, the number, order, and grouping of output slots must exactly match the schema.

\paragraph{Formula hypergraph.}
For calculation problems, we extract formula traces from reference solutions and organize them into a formula hypergraph. Each formula instance is represented as a hyperedge connecting input variables, intermediate quantities, and answer targets. Formula records also store slot notes, formula meanings, variable definitions, and source metadata when available. This representation allows retrieval at the level of solution operations rather than isolated equations.

\paragraph{Question-only formula retrieval.}
In the non-oracle retrieval setting, the model receives formula candidates retrieved using only the current question text. Each candidate target document is built from the corresponding question, slot notes, and formula strings. We rank candidate targets using lexical overlap:
\begin{equation}
s(q,d)
=
\frac{|T(q)\cap T(d)|}{|T(q)\cup T(d)|}
+
0.01|T(q)\cap T(d)|,
\end{equation}
where \(T(q)\) and \(T(d)\) are token sets for the query and candidate document. Retrieved formulas are presented as candidate solution templates, not as ground truth. The prompt explicitly instructs the model to verify whether each formula matches the current problem, variables, units, and rounding constraints before using it.

\paragraph{Oracle formulas and distractors.}
We also evaluate controlled formula assistance. In the oracle-formula setting, the model receives the ground-truth formula traces for the current problem as method guidance, but it must still map variables, compute values, handle units, and satisfy the answer schema. In the oracle-with-distractors setting, the ground-truth formula set is hidden among distractor formula sets. Exactly one candidate set is correct, but the correct label is not revealed. This setting separates formula-plan recognition from numerical execution.

\paragraph{Two-round code augmentation.}
For tool-augmented inference, we use a two-round protocol. In the first round, the model is required to generate Python code for solving the problem. We execute the generated code locally in a temporary working directory with a timeout, and then append the execution output to the conversation. In the second round, the model uses the execution result to produce the final answer in the same \texttt{<answer>} format as zero-shot inference. The prompt instructs the model to use only the Python standard library and avoid interactive input, network calls, and file writes. We also evaluate an oracle-formula code setting, where the ground-truth formulas are provided before code generation.

\section{Further Difficulty Analysis}

\begin{figure*}[!htbp]
    \centering
    \includegraphics[width=\linewidth]{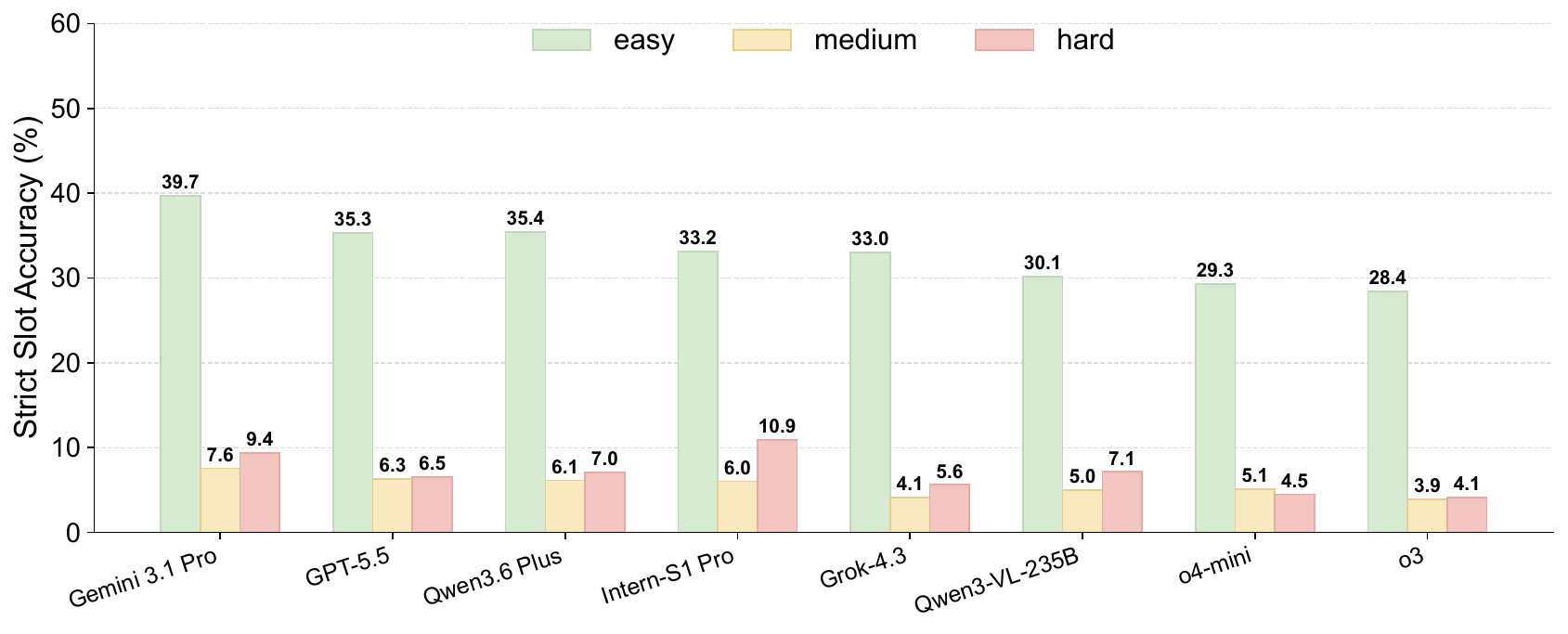}
    \caption{Model performance vs. Difficulty Levels (Overall)}
    \label{fig:calculation-slot-accuracy-2}
\end{figure*}

\subsection{Overall Difficulty-wise CAL Analysis}
\label{app:cal_difficulty_overall}

Figure~\ref{fig:calculation-slot-accuracy-2} summarizes the overall CAL strict-slot accuracy under different difficulty levels.
The most salient observation is the large gap between easy examples and non-easy examples.
Across all evaluated models, easy accuracy remains substantially higher, ranging from 28.4\% to 39.7\%.
In contrast, medium and hard accuracy are much lower, with most scores below 8\% and all scores below 11\%.
For example, Gemini 3.1 Pro achieves the highest easy accuracy of 39.7\%, but its medium and hard accuracies drop to 7.6\% and 9.4\%, respectively.
A similar pattern appears for GPT-5.5 and Qwen3.6 Plus, whose easy accuracies are above 35\%, while their medium and hard accuracies remain around 6--7\%.
This suggests that current MLLMs can handle simpler CAL instances to some extent, but their strict slot-level reliability drops sharply once examples require more complex grounding, finer-grained slot selection, or less explicit evidence.

A second observation is that the difficulty trend is not strictly monotonic between medium and hard examples.
For most models, hard accuracy is slightly higher than medium accuracy; Intern-S1 Pro is the clearest case, obtaining 6.0\% on medium examples but 10.9\% on hard examples.
This does not imply that hard examples are intrinsically easier.
Rather, it suggests that strict-slot CAL difficulty is affected not only by the nominal difficulty label, but also by the distribution of categories, slot types, and textual cues within each split.
Some hard examples may contain more distinctive domain signals, while medium examples may involve more ambiguous slot boundaries or require more precise category-specific normalization.
Therefore, the difficulty-wise results indicate that CAL primarily exposes a robustness gap in structured slot grounding, rather than a simple monotonic degradation from easy to hard.
\section{Annotation, Human Reference, and Cost Details}

To ensure domain coverage and annotation quality, we involved a mixed team of materials-science annotators and validators, including 8 PhD students, 60 postgraduate and undergraduate students, and 5 materials-domain professors. Undergraduate and postgraduate annotators participated in problem collection, formatting, and preliminary consistency checks, while PhD students and professors were responsible for domain-level verification, formula validation, solution checking, and resolving ambiguous cases. This multi-stage process allowed us to combine scalable data construction with expert oversight.

\paragraph{Constrained Human Baseline.}
\label{app:human_baseline}
To provide a human reference point, we recruited two PhD students with materials-science backgrounds and asked them to solve a randomly sampled subset of \textsc{OmniMatBench}. The sampled questions were drawn across both QA and CAL tasks and were balanced across the four coarse-grained materials-science groups whenever possible. Participants were given the same problem statements, multimodal inputs, and answer-format requirements as the evaluated models, but were not allowed to use LLM assistance. For QA questions, their responses were scored using the same expert-curated key-point rubrics as model responses. For CAL questions, their answers were parsed into the same structured answer slots and evaluated with the same strict exact-slot accuracy. We report the average score across the two participants as the human reference baseline.

This reference is intended to contextualize the strictness of the evaluation protocol, not to estimate an upper bound on expert materials-science performance. Participants solved the problems under constrained conditions without LLM assistance or external computational tools, and their answers were evaluated with the same strict slot and formatting requirements used for model outputs. Therefore, a model score above this reference should not be interpreted as evidence that the model surpasses human experts.
\begin{table}[h]
\centering
\small
\begin{tabularx}{\linewidth}{Xc}
\toprule
Evaluation Metric & Mean Absolute Error \\
\midrule
Recall (Completeness) & 0.126 \\
Precision (Correctness) & 0.153 \\
F$_1$-Score & 0.122 \\
\bottomrule
\end{tabularx}
\caption{Agreement between Gemini 2.5 Flash and human experts on sampled QA responses.}
\label{tab:evaluator-validation}
\end{table}

\paragraph{Inference Cost Reference.}
To contextualize the computational cost of our model evaluation, we report approximate public reference prices for representative models in Table~\ref{tab:appendix-model-cost}. Prices are listed in USD per one million input and output tokens. The actual billing rate may depend on the selected routing group and platform-side pricing. We therefore report only models for which public reference prices are available, and treat the table as a cost reference rather than an audited billing record.

\begin{table}[!htbp]
\centering

\caption{Approximate public reference prices for representative models used in our CAL experiments. Prices are in USD per one million tokens.}
\label{tab:appendix-model-cost}
\setlength{\tabcolsep}{3.2pt}
\begin{tabular}{lcc}
\toprule
Model & Input & Output \\
\midrule
GPT-5.5 & 5.00 & 30.00 \\
Claude Opus 4.7 & 5.00 & 25.00 \\
Gemini 3.1 Pro & 2.00 & 12.00 \\
o3 & 2.00 & 8.00 \\
o4-mini & 1.10 & 4.40 \\
Grok-4.3 & 1.25 & 2.50 \\
\bottomrule
\end{tabular}
\end{table}

\begin{table}[!htbp]
\centering
\caption{Small-model performance on OmniMatBench. 
Scores are macro-averaged over four representative materials-science categories.}
\label{tab:small_model_cal_ablation}
\scriptsize
\setlength{\tabcolsep}{4.5pt}
\resizebox{\columnwidth}{!}{
\begin{tabular}{llc}
\toprule
\textbf{Model} & \textbf{Setting} & \textbf{Macro Avg.} \\
\midrule
LLaMA-3.2-11B-Vision-Instruct 
& Instruct & 1.27 \\

Qwen3-VL-8B-Instruct 
& Instruct & 12.45 \\

Qwen3-VL-8B-Thinking 
& Thinking & \textbf{18.00} \\
\bottomrule
\end{tabular}
}

\vspace{2pt}
\footnotesize
\textit{Note:} Scores are macro-averaged over over sampled subset. 
\end{table}


\section{Prompts}
\subsection{Prompt Templates}
\label{app:prompts}

This section lists the main prompt templates used in our experiments. Dynamic
fields such as the question text, image input, formula context, execution
results, and answer-format schema are inserted at runtime.

\begin{figure}[!htbp]
\centering
\begin{tcolorbox}[
    enhanced,
    colback=white,
    colframe=cvprblue,
    title={\textbf{Zero-shot Inference Prompt}},
    width=0.98\linewidth,
    arc=2pt, boxrule=0.8pt,
    left=6pt, right=6pt, top=6pt, bottom=6pt,
    fonttitle=\bfseries,
    fontupper=\scriptsize\sffamily
]
\textbf{Role:} You are a materials science expert. Your task is to provide a clear and accurate answer.

\textbf{Reasoning and Answer Format:} Put internal reasoning inside \texttt{<think>...</think>} and put only the final answer content inside \texttt{<answer>...</answer>}. Do not include derivation, explanation, labels, or extra commentary inside \texttt{<answer>}.

\textbf{Slot Constraint:} If \texttt{final\_answer\_format} is provided, follow its grouping, ordering, and slot count exactly. Replace each empty slot with one final answer string containing only the final number, formula, or expression.

\textbf{User Input:} The user message contains \texttt{\{question\_text\}}, optional image or CSV input, and optionally \texttt{final\_answer\_format: \{final\_answer\_format\}}.
\end{tcolorbox}
\caption{The full prompt template for zero-shot inference.}
\label{fig:prompt_zeroshot_full}
\end{figure}

\begin{figure}[!htbp]
\centering
\begin{tcolorbox}[
    enhanced,
    colback=white,
    colframe=cvprblue,
    title={\textbf{Formula-grounded / Formula-retrieval Prompt}},
    width=0.98\linewidth,
    arc=2pt, boxrule=0.8pt,
    left=6pt, right=6pt, top=6pt, bottom=6pt,
    fonttitle=\bfseries,
    fontupper=\scriptsize\sffamily
]
\textbf{Base Prompt:} The base zero-shot prompt is used, with an additional formula-assistance block appended to the user message.

\textbf{Formula Assistance:} The model may use \texttt{\{formula\_slots\_or\_retrieved\_formula\_candidates\}} to solve the problem, but should select and apply only formulas relevant to the question.

\textbf{Candidate Selection Rule:} When candidate formula groups are provided, the set contains one correct formula plan and several distractor plans. The model must identify the applicable plan from the question context, avoid assuming that every provided formula is relevant, and still follow \texttt{final\_answer\_format} exactly.
\end{tcolorbox}
\caption{The full prompt template for formula-grounded and formula-retrieval inference.}
\label{fig:prompt_formula_grounded_full}
\end{figure}

\begin{figure}[!htbp]
\centering
\begin{tcolorbox}[
    enhanced,
    colback=white,
    colframe=cvprblue,
    title={\textbf{Formula-selection-only Prompt}},
    width=0.98\linewidth,
    arc=2pt, boxrule=0.8pt,
    left=6pt, right=6pt, top=6pt, bottom=6pt,
    fonttitle=\bfseries,
    fontupper=\scriptsize\sffamily
]
\textbf{Role:} You are a materials science expert. Your task is to select the formula candidate group that should be used to solve the problem.

\textbf{Input:} The user message contains \texttt{\{question\_text\}} and a list of formula candidate groups, \texttt{\{candidate\_formula\_plans\}}.

\textbf{Output Rule:} Return only the selected candidate label inside \texttt{<answer></answer>}. Do not solve the full numerical problem and do not include explanation inside \texttt{<answer>}.

\textbf{Required Format:} Output \texttt{<think>} with a brief justification of which candidate group matches the problem, followed by \texttt{<answer>["\{candidate\_label\}"]</answer>}.
\end{tcolorbox}
\caption{The full prompt template for formula-selection-only evaluation.}
\label{fig:prompt_formula_selection_full}
\end{figure}

\begin{figure}[!htbp]
\centering
\begin{tcolorbox}[
    enhanced,
    colback=white,
    colframe=cvprblue,
    title={\textbf{Tool-augmented Code Prompt}},
    width=0.98\linewidth,
    arc=2pt, boxrule=0.8pt,
    left=6pt, right=6pt, top=6pt, bottom=6pt,
    fonttitle=\bfseries,
    fontupper=\scriptsize\sffamily
]
\textbf{Role:} You are a materials science expert. Solve the question accurately.

\textbf{Two-phase Behavior:} In the first response, the model must output exactly one clean Python code block in triple backticks, must not output the final answer, and must not skip code generation. If the user later provides execution results, the model must use those results to produce the final answer.

\textbf{Code Constraints:} The code must use only the Python standard library, be self-contained and deterministic, print key computed values clearly, and avoid \texttt{input()}, network calls, file writes, or external packages.

\textbf{User Input and Final Answer:} The user message contains \texttt{\{question\_text\}}, optional image or CSV input, and possibly \texttt{final\_answer\_format: \{final\_answer\_format\}}. The second-round response must include reasoning in \texttt{<think>...</think>} and only the final answer content in \texttt{<answer>...</answer>}, following all slot grouping and ordering constraints.
\end{tcolorbox}
\caption{The full prompt template for tool-augmented code generation.}
\label{fig:prompt_tool_code_full}
\end{figure}

\begin{figure}[!htbp]
\centering
\begin{tcolorbox}[
    enhanced,
    colback=white,
    colframe=cvprblue,
    title={\textbf{Tool Execution Feedback Prompt}},
    width=0.98\linewidth,
    arc=2pt, boxrule=0.8pt,
    left=6pt, right=6pt, top=6pt, bottom=6pt,
    fonttitle=\bfseries,
    fontupper=\scriptsize\sffamily
]
\textbf{Execution Feedback:} Here are the execution results of your Python code from the previous turn: \texttt{\{code\_executed\}}.

\textbf{Final Answer Instruction:} Now provide your final answer for the original question. Your response must include both \texttt{<think>...</think>} and \texttt{<answer>...</answer>}. Inside \texttt{<answer>}, output only the final answer content.
\end{tcolorbox}
\caption{The full prompt template for providing code execution feedback to the model.}
\label{fig:prompt_tool_feedback_full}
\end{figure}


\section{Qualitative Error Case Studies}
\label{app:qualitative-error-cases-raw}

We present raw qualitative examples from \texttt{gpt-5.5}, with one
representative multimodal case for each error type used in our analysis:
\texttt{calculation\_error}, \texttt{formula\_error},
\texttt{conceptual\_error}, \texttt{unit\_conversion\_error},
\texttt{rounding\_error}, \texttt{image\_understanding\_error}, and
\texttt{format\_error}. For each case, we show the benchmark image, the
original question text, the original ground-truth answer field, the model's
reasoning trace, the model's raw answer block, and a concise manual diagnosis
based on the trace and final answer. All examples are drawn directly from
\texttt{result/gpt-5.5} and matched to the current \texttt{0513\_CAL} source by
exact question text.

\subsection{Case 1: Calculation error: incorrect fatigue-life computation}

\noindent
\begin{minipage}{\columnwidth}
\centering
\includegraphics[width=\columnwidth]{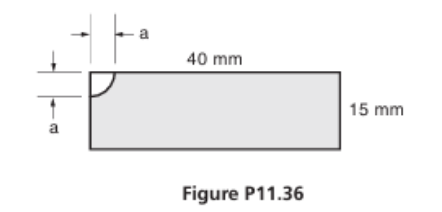}
\end{minipage}

\vspace{0.5em}

\noindent\textbf{Benchmark ID:} CAL 05 / id 031\\
\textbf{Error label:} \texttt{calculation\_error}\\
\textbf{Manual diagnosis:} The trace applies a damage-tolerance crack-growth calculation, but the final cycle count is far from the reference value, indicating a numerical/substitution error rather than a missing visual input.

\vspace{0.75em}
\noindent\textbf{Original Question} An aircraft structural member made of 7075-T6 aluminum has a cross section as shown in Fig. A quarter circular corner crack of size a = 0.5mm is present, and the member is subjected to a tension stress S. How many cycles between Smax = 336 and Smin =-68MPa can be applied before failure is expected? (Comment: It is reasonable to assume that an approximately quarter-circular crack shape is maintained as the crack grows. Also, conservatively approximate the fully plastic limit force by assuming that the crack of depth a extends uniformly across the full 15mm thickness of the member.)(For this question Unit: $\text{cycles}$ rounded to integer)

\noindent\textbf{Original Ground Truth Answer} $N_{if} = 4240$ cycles.

\noindent\textbf{GPT-5.5 Reasoning Trace}

Using the standard damage-tolerance calculation for a quarter-circular corner crack in 7075-T6 aluminum, the crack is grown from the initial 0.5 mm size to the critical size governed by fracture before net-section plastic collapse. Integrating the fatigue crack-growth law over that interval gives approximately $1.2*10^4$ cycles.

\noindent\textbf{Parsed GPT-5.5 Answer}
["12000"]

\subsection{Case 2: Formula error: including pore-air density in open-cell foam calculation}

\noindent
\begin{minipage}{\columnwidth}
\centering
\includegraphics[width=\columnwidth]{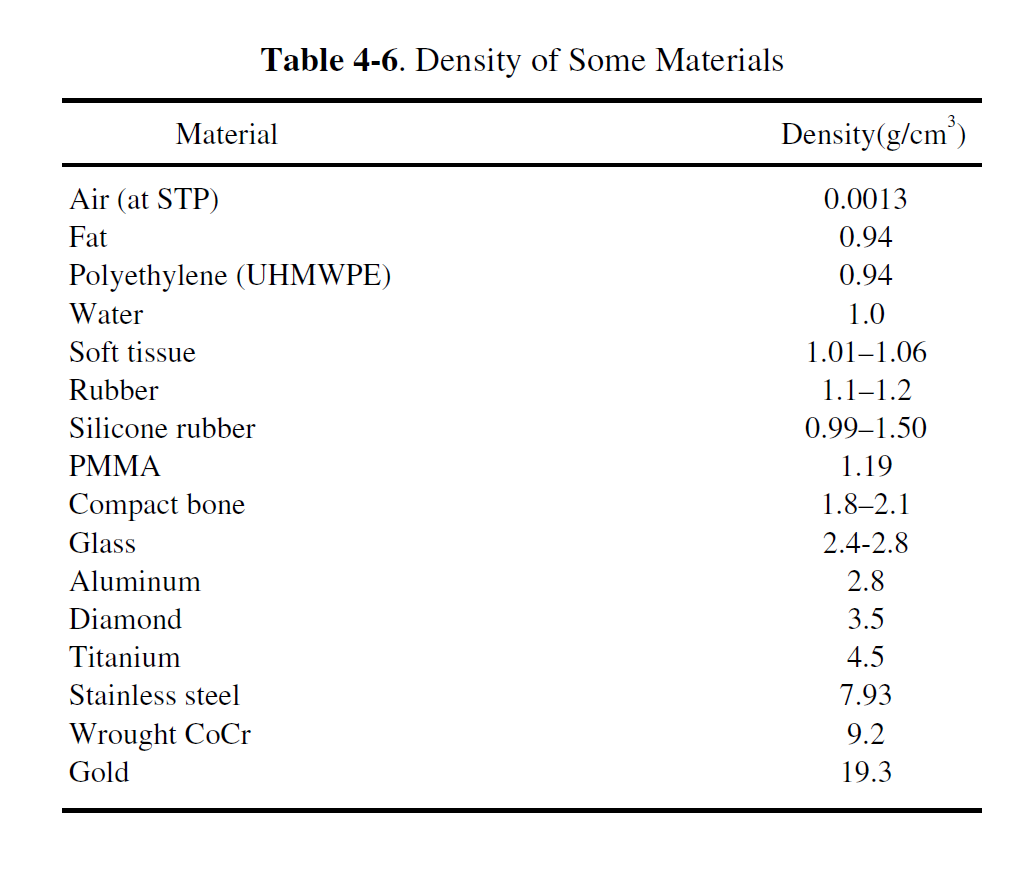}
\end{minipage}

\vspace{0.5em}
\noindent\textbf{Benchmark ID:} CAL 16 / id 024\\
\textbf{Error label:} \texttt{formula\_error}\\
\textbf{Manual diagnosis:} The trace explicitly includes the density of air in the pores when computing apparent density and sheet weight, whereas the reference solution uses only the solid polyurethane volume fraction.

\vspace{0.75em}
\noindent\textbf{Original Question}
\noindent \textbf{Example 4.5} \\ A polyurethane open cell foam is to be used for a lining for an artificial leg. The solid volume fraction is 4\%, the pores are 0.5 mm in diameter and the solid polyurethane has a density of 0.9 g/cm$^3$. Determine the porosity and apparent density of the material, and the weight of a sheet 200 mm by 200 mm by 1 cm thick.Round to 3 decimal places.

\noindent\textbf{Original Ground Truth Answer}

\begin{table}[htbp]
        \centering
        \caption{Table 4-6. Density of Some Materials}
        \begin{tabular}{|lc|}
                \hline
                Material & Density (g/cm$^3$) \\
                \hline\hline
                Air (at STP) & 0.0013 \\
                Fat & 0.94 \\
                Polyethylene (UHMWPE) & 0.94 \\
                Water & 1.0 \\
                Soft tissue & 1.01--1.06 \\
                Rubber & 1.1--1.2 \\
                Silicone rubber & 0.99--1.50 \\
                PMMA & 1.19 \\
                Compact bone & 1.8--2.1 \\
                Glass & 2.4--2.8 \\
                Aluminum & 2.8 \\
                Diamond & 3.5 \\
                Titanium & 4.5 \\
                Stainless steel & 7.93 \\
                Wrought CoCr & 9.2 \\
                Gold & 19.3 \\
                \hline
        \end{tabular}
\end{table}

$
\text{Porosity } P = 1 - V_s, \ \text{so } P = 1 - 0.04 = 0.96, \ \text{or } 96\% \ \text{porosity}.
$
$
\text{Apparent density} = \rho_{\text{app}} = V_s \rho_{\text{true}} = 0.04 \left[ 0.9\ \text{g/cm}^3 \right] = \underline{0.036\ \text{g/cm}^3}.
$
$
\text{Mass} = V \rho_{\text{app}} = 20 \times 20 \times 1\ (\text{cm}^3)\ 0.036\ \text{g/cm}^3 = \underline{14.4\ \text{g}}, \ \text{so weight} = \underline{0.032\ \text{lb}}.
$
The weight of the foam lining is much less than that of an artificial leg. In some cases, the effect of implant density is important. For example, intraocular lenses made of PMMA are denser than the soft tissues of the eye, as shown in Table 4-6. During rapid eye movement, the lens is accelerated by the structures to which it is attached, e.g., the iris. By contrast, the natural eye lens is essentially neutrally buoyant. Damage to these eye structures by intraocular lenses has been observed clinically. Lenses of a silicone rubber composition of lower density are currently under investigation; these offer the added benefit of being softer, and hence easier to insert through a small incision. Another example of the density of implants is augmentation mammoplasty, in which the added weight of the implant can be problematical.

\noindent\textbf{Ground Truth Answer Slots}
[
  "0.96",
  "0.036",
  "0.032"
]

\noindent\textbf{GPT-5.5 Reasoning Trace}

Solid volume fraction = 0.04, so porosity = 1 - 0.04 = 0.96. For open-cell foam, apparent density includes solid polyurethane plus air in pores: ρ = 0.04(0.9) + 0.96(0.0013) = 0.037248 g/cm³. Sheet volume = 20 cm × 20 cm × 1 cm = 400 cm³, so mass/weight = 0.037248 × 400 = 14.8992 g. Rounded to 3 decimals.

\noindent\textbf{Parsed GPT-5.5 Answer}
[
  "0.960",
  "0.037",
  "14.899"
]

\subsection{Case 3: Conceptual error: wrong sign convention 
for compressive yielding}

\noindent
\begin{minipage}{\columnwidth}
\centering
\includegraphics[width=\columnwidth]{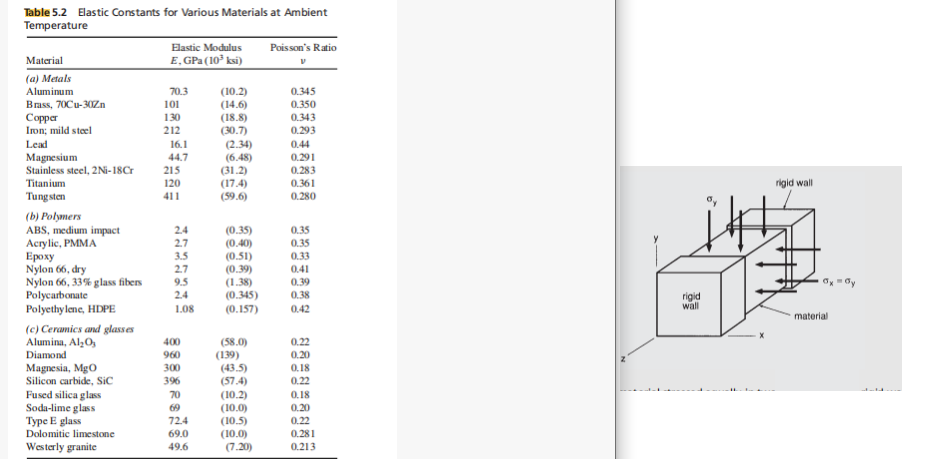}
\end{minipage}

\vspace{0.5em}
\noindent \textbf{Benchmark ID:} CAL 05 / id 001\\
\textbf{Error label:} \texttt{conceptual\_error}\\
\textbf{Manual diagnosis:} The trace derives the constrained stress relation but reports the yielding stress as positive/tensile, missing the compressive sign convention required by the problem.

\vspace{0.75em}
\noindent\textbf{Original Question}

A block of material is subjected to equal compressive stresses in the x- and y-directions, and it is confined by a rigid die so that it cannot deform in the z-direction, as shown in Fig. Assume that there is no friction against the die and also that the material behaves in an elastic, perfectly plastic manner, with uniaxial yield strength $\sigma_o$. (a) Determine the stress $\sigma_x = \sigma_y$ necessary to cause yielding, expressing this as a function of $\sigma_o$ and elastic constants of the material. (b) What is the value of $\sigma_y$ at yielding if the material is an aluminum alloy with uniaxial yield strength $\sigma_o = 300 \, \text{MPa}$ and elastic constants as in Table 5.2? ---(question b Unite:$\text{ MPa}$ round to one decimal place)

\noindent\textbf{Original Ground Truth Answer}
(a) Apply Hooke’s law for the $z$-direction,letting $\sigma_x = \sigma_y$, and noting that preventing deformation in the $z$-direction requires that the strain in that direction be zero ($\epsilon_z = 0$):
$$ \epsilon_z = \frac{1}{E} [\sigma_z - \nu(\sigma_x + \sigma_y)] ,$$ $$\quad 0 = \frac{1}{E} [\sigma_z - \nu(\sigma_y + \sigma_y)], \quad \sigma_z = 2\nu\sigma_y $$
Here, solving the second expression for $\sigma_z$ gives the third. Since there are no shear stresses, the $x$-$y$-$z$ axes are also the principal axes, 1-2-3, and the principal normal stresses are
$$ \sigma_1 = \sigma_x = \sigma_y, \quad \sigma_2 = \sigma_y, \quad \sigma_3 = \sigma_z = 2\nu\sigma_y $$

The effective stress for the octahedral shear criterion is
$$ \bar{\sigma}_H = \frac{1}{\sqrt{2}}\sqrt{(\sigma_1 - \sigma_2)^2 + (\sigma_2 - \sigma_3)^2 + (\sigma_3 - \sigma_1)^2} $$
$$ \bar{\sigma}_H = \frac{1}{\sqrt{2}}\sqrt{(\sigma_y - \sigma_y)^2 + (\sigma_y - 2\nu\sigma_y)^2 + (2\nu\sigma_y - \sigma_y)^2} $$
$$ \bar{\sigma}_H = \sigma_y(1 - 2\nu) $$
Since $\bar{\sigma}_H = \sigma_o$ at the point of yielding, the desired result is
$$ \sigma_y = \frac{\sigma_o}{1 - 2\nu}  $$

(b) For the aluminum alloy with uniaxial yield strength $\sigma_o = 300 \text{ MPa}$, assume that $\sigma_o = -300 \text{ MPa}$ applies for uniaxial compression. Substituting this and $\nu = 0.345$ from Table 5.2, the stress to cause yielding is
$$ \sigma_y = \frac{-300 \text{ MPa}}{1 - 2(0.345)} = -967.7 \text{ MPa} $$

\noindent\textbf{Ground Truth Answer Slots}
[
  $\sigma\_y = \frac{\sigma_o}{1 - 2\nu}$,
  "-967.7"
]

\noindent\textbf{GPT-5.5 Reasoning Trace}
Used the zero z-strain condition to relate the die reaction stress to the applied equal compressive stresses, then applied the yield condition.

\noindent\textbf{Parsed GPT-5.5 Answer}

[
  $\sigma\_y = \frac{\sigma_o}{1 - 2\nu}$,
  "967.7"
]

\subsection{Case 4: Unit conversion error: mechanics quantities reported at the wrong scale}

\noindent
\begin{minipage}{\columnwidth}
\centering
\includegraphics[width=\columnwidth]{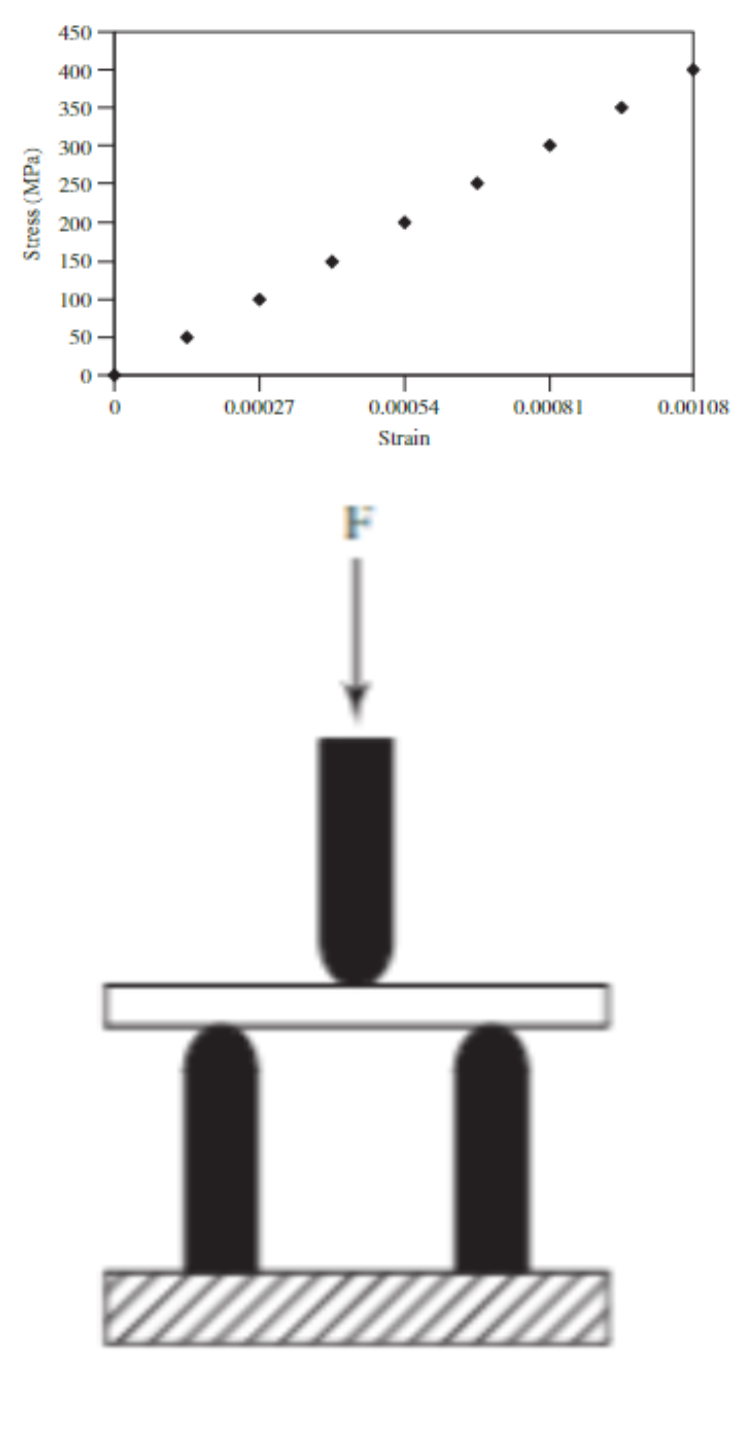}
\end{minipage}

\vspace{0.5em}
\noindent\textbf{Benchmark ID:} CAL 16 / id 044\\
\textbf{Error label:} \texttt{unit\_conversion\_error}\\
\textbf{Manual diagnosis:} The trace uses the expected slope and bending relations in form, but the outputs are in inconsistent scales/units, e.g., 370370 instead of 370 GPa and 12566.4 instead of 4.2 N.

\vspace{0.75em}
\noindent\textbf{Original Question}

\noindent \textbf{EXAMPLE PROBLEM 4.6} The following is a stress-strain curve resulting from a three-point bending test on an alumina (ceramic) test sample: 
\begin{tikzpicture}
\begin{axis}[
    width=8cm,
    height=6cm,
    xlabel={Strain},
    ylabel={Stress (MPa)},
    xmin=0, xmax=0.0011,
    ymin=0, ymax=450,
    xtick={0, 0.00027, 0.00054, 0.00081, 0.00108},
    ytick={0, 50, 100, 150, 200, 250, 300, 350, 400, 450},
    tick label style={font=\small, black},
    label style={font=\small, black},
    only marks,
    mark=diamond*,
    mark size=3,
    every mark/.append style={fill=black, draw=black},
    axis lines=box,
    tick style={thick, black},
    axis line style={thick, black},
    scaled x ticks=false,
    xticklabel style={
        /pgf/number format/fixed,
        /pgf/number format/precision=6
    },
    yticklabel style={
        /pgf/number format/fixed,
        /pgf/number format/precision=0
    }
]
\addplot coordinates {
    (0, 0)
    (0.000135, 50)
    (0.00027, 100)
    (0.000405, 150)
    (0.00054, 200)
    (0.000675, 250)
    (0.00081, 300)
    (0.000945, 350)
    (0.00108, 400)
};
\end{axis}
\end{tikzpicture}

\begin{enumerate}
    \item[(a)] Calculate the modulus of elasticity of the alumina sample. Round to integer.
    \item[(b)] Using this information, what is the modulus of rupture? Round to integer.
    \item[(c)] The sample is a cylindrical specimen with a radius of 1 cm and a distance of 10 cm between the lower supports. What force was necessary to cause this fracture? Round to 1 decimal place.
    \item[(d)] Given the following diagram of the three-point bending test of a rectangular sample, which portion of the sample is in compression and which portion is in tension?
\end{enumerate}

\noindent\textbf{Original Ground Truth Answer}

\noindent\textbf{Solution:}\begin{enumerate} \item[(a)] The modulus of elasticity can be determined from the slope of the linear portion of the stress-strain curve: \[ E = \frac{\Delta \sigma}{\Delta \varepsilon} = \frac{400 - 0\ \text{MPa}}{0.00108 - 0} = {370\ \text{GPa}} \]
 \item[(b)] The modulus of rupture is the stress at the point of fracture, around \({400\ \text{MPa}}\).
 \item[(c)] Solve equation 4.25 for \(F_f\) and determine the force at fracture: \[ F_f = \frac{\sigma_{\text{max}} \pi R^3}{3L} = \frac{400000 \pi (0.01)^3}{3 \times 0.1} \approx {4.2\ \text{N}} \]
 \item[(d)] The portion of the sample at the top facing the indenting attachment is in compression, while the portion of the sample at the bottom facing the two supporting points is in tension under the load.\end{enumerate}

\noindent\textbf{Ground Truth Answer Slots}
[
  "370",
  "400",
  "4.2",
  "The portion of the sample at the top facing the indenting attachment is in compression, while the portion of the sample at the bottom facing the two supporting points is in tension under the load."
]

\noindent\textbf{GPT-5.5 Reasoning Trace}
Computed using the stress–strain slope, fracture stress, and circular three-point bending relation.

\noindent\textbf{Parsed GPT-5.5 Answer}
[
  "370370",
  "400",
  "12566.4",
  "Top in compression; bottom in tension"
]

\subsection{Case 5: Rounding error: smectic order-parameter ratio differs in the last digit}

\noindent
\begin{minipage}{\columnwidth}
\centering
\includegraphics[width=\columnwidth]{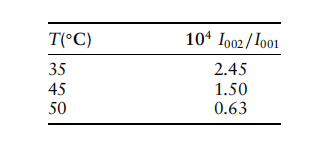}
\end{minipage}

\vspace{0.5em}
\noindent\textbf{Benchmark ID:} CAL 04 / id 027\\
\textbf{Error label:} \texttt{rounding\_error}\\
\textbf{Manual diagnosis:} The trace follows the intended square-root intensity-ratio calculation, but the first reported ratio is 0.0157 rather than the reference 0.0156.

\vspace{0.75em}
\noindent\textbf{Original Question}

The coefficients $\rho_n$ in Eq. (\[\rho(z) = \rho_0 + \sum_n \rho_n \cos(nq^*z - \Phi_n)\]) are the smectic order parameters and describe the sharpness of the layering. It is possible to obtain estimates of the ratio of these order parameters from the intensities of peaks in x-ray diffraction patterns. Specifically,

\[\frac{I_{00l}}{I_{001}} \approx \left( \frac{\rho_l}{\rho_1} \right)^2\]

where $I_{00l}$ is the intensity of the $l$th-order Bragg reflection. For a molecule with the structure of HOAB but with the alkoxy groups replaced by alkyl groups (in Fig), the following values were reported from x-ray diffraction experiments in the smectic A phase: Estimate the ratio $\rho_2/\rho_1$ at these three temperatures. What does this typical data tell you about the structure of the smectic A phase and its temperature dependence? Why is the above equation an approximation, and does it over- or underestimate the true ratio $\rho_2/\rho_1$? (For this question round to four decimal places)

\noindent\textbf{Original Ground Truth Answer}

T = 35℃, $\rho_2$/$\rho_1$ = 0.0156；T = 45℃, $\rho_2$/$\rho_1$ = 0.0122；T = 50℃, $\rho_2$/$\rho_1$ = 0.0079. Underestimate because the molecular form factor is neglected.

\noindent\textbf{Ground Truth Answer Slots}

[
  "0.0156",
  "0.0122",
  "0.0079",
  "Underestimate"
]

\noindent\textbf{GPT-5.5 Reasoning Trace}

Use $\rho_2$/$\rho_1$ $\approx$ $\sqrt{(I_{002}/I_{001})}$, with tabulated values multiplied by $10^{-4}$. This gives 0.01$\sqrt{2.45}$, 0.01$\sqrt{1.50}$, and 0.01$\sqrt{0.63}$. The small, decreasing second harmonic indicates nearly sinusoidal smectic-A layering that becomes less sharp at higher temperature. The intensity relation neglects molecular/atomic form-factor effects, which generally reduce higher-order reflections, so it underestimates the true ratio.

\noindent\textbf{Parsed GPT-5.5 Answer}
[
  "0.0157",
  "0.0122",
  "0.0079",
  "nearly sinusoidal smectic-A layering with decreasing sharpness as temperature increases; approximation neglects form-factor effects and underestimates the true $\rho_2$/$\rho_1$"
]

\subsection{Case 6: Image understanding error: wrong zero crossing in a zeta-potential plot}

\noindent
\begin{minipage}{\columnwidth}
\centering
\includegraphics[width=\columnwidth]{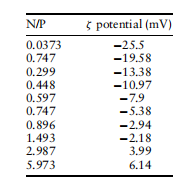}
\end{minipage}

\vspace{0.5em}
\noindent\textbf{Benchmark ID:} CAL 04 / id 025\\
\textbf{Error label:} \texttt{image\_understanding\_error}\\
\textbf{Manual diagnosis:} The trace states that the zero crossing is approximately N/P = 2, but the visual/reference answer indicates charge neutralization near N/P = 1.

\vspace{0.75em}
\noindent\textbf{Original Question}

A solution of DNA was titrated with a cationic block copolymer and the zeta potential ($\\zeta$) was measured, with the following results:Here $N/P$ is the mole ratio of nitrogens in the cationic block copolymer to the number of phosphate groups in DNA. [Data from J. F. Tan *et al. Langmuir* 22, 3744 (2006)].Determine the $N/P$ ratio at which charge neutralization occurs. Comment on the relationship of this data to the Manning condensation point."(For this question Keep as integer)

\noindent\textbf{Original Ground Truth Answer}

$N/P \approx 1$

\noindent\textbf{Ground Truth Answer Slots}
[
  "1"
]

\noindent\textbf{GPT-5.5 Reasoning Trace}

Interpolated zero crossing gives approximately 2; rounded to an integer per the constraint.

\noindent\textbf{Parsed GPT-5.5 Answer}
[
  "2"
]

\subsection{Case 7: Format error: equivalent duration in a nonmatching textual format}

\noindent
\begin{minipage}{\columnwidth}
\centering
\includegraphics[width=\columnwidth]{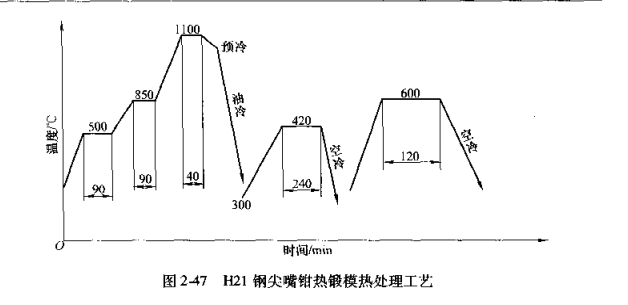}
\end{minipage}

\vspace{0.5em}
\noindent \textbf{Benchmark ID:} CAL 01 / id 033\\
\textbf{Error label:} \texttt{format\_error}\\
\textbf{Manual diagnosis:} The trace computes the correct duration, 580 minutes = 9 hours 40 minutes, but the final answer uses the abbreviated form 9 h 40 min instead of the expected textual slot.

\vspace{0.75em}
\noindent\textbf{Original Question}
A hot forging die for pointed nose pliers made of H21 steel, with dimensions of 75mm x 75mm x 105mm, is used on a 3000kN friction press. The forging material is 45 steel or 40Cr steel, and the forging frequency is 6~7 pieces per minute. The average service life of the die treated by conventional technology is only about 4000 pieces, and the failure modes are cracking or cavity deformation and collapse. However, by adopting isothermal quenching and heat treatment to control the hardened layer depth on the die surface, the service life can exceed 20000 pieces. The heat treatment processes are as follows: 1. Salt bath isothermal quenching process (Process 1): - Two-stage preheating at 600°C and 850°C; - Heating at 1050~1060°C, pre-cooling to 950°C, then quenching into 280°C nitrate salt for 5 minutes; - Immediately transferring to 380°C nitrate salt for isothermal holding for 3~4 hours, followed by air cooling; - Tempering twice at 360°C for 4 hours each time. 2. Box furnace high-temperature short-time heating + controlled cooling composite process (Process 2): - Preheating at 500°C for 90 minutes -> preheating at 850°C for 90 minutes -> heating at 1100°C for 40 minutes; - Pre-cooling, then oil cooling to 300°C -> holding at 420°C for 240 minutes -> air cooling; - Tempering at 600°C for 120 minutes, followed by air cooling. Given that the effective thickness of the hot forging die for pointed nose pliers is 75mm, calculate the total process time from preheating to tempering completion in Process 2, and write out the calculation process. Express the result in minutes as an integer, and also give the equivalent time in hours and minutes format.

\noindent\textbf{Original Ground Truth Answer}

1. Clarify the duration of each stage (strictly based on Figure 2-47 and process description) - Preheating stage: 500°C preheating: 90 min, 850°C preheating: 90 min, Total preheating time: \(90 + 90 = 180\ \text{min}\); - High-temperature heating stage: 1100°C heating and holding: 40 min (clearly marked as "40" in the figure); - Quenching and isothermal holding stage: Pre-cooling + oil cooling to 300°C: No clearly marked duration, not included in fixed holding time, 420°C isothermal holding: 240 min (clearly marked as "240" in the figure); - Tempering stage: 600°C tempering and holding: 120 min (clearly marked as "120" in the figure); 2. Total process time calculation: \(t_{\text{total}} = 180\ \text{min} + 40\ \text{min} + 240\ \text{min} + 120\ \text{min} = 580\ \text{min}\); 3. Time unit conversion: \(580\ \text{min} = 9\ \text{hours}\ 40\ \text{minutes}\)

\noindent\textbf{Ground Truth Answer Slots}
[
  "580",
  "9 hours 40 minutes"
]

\noindent\textbf{GPT-5.5 Reasoning Trace}
Total specified durations in Process 2: 90 + 90 + 40 + 240 + 120 = 580 min, which equals 9 h 40 min.

\noindent\textbf{Parsed GPT-5.5 Answer}
[
  "580",
  "9 h 40 min"
]


\subsection{Case 8: Wrong formula use in open-cell foam density}

\noindent
\begin{minipage}{\columnwidth}
\centering
\includegraphics[width=0.82\columnwidth]{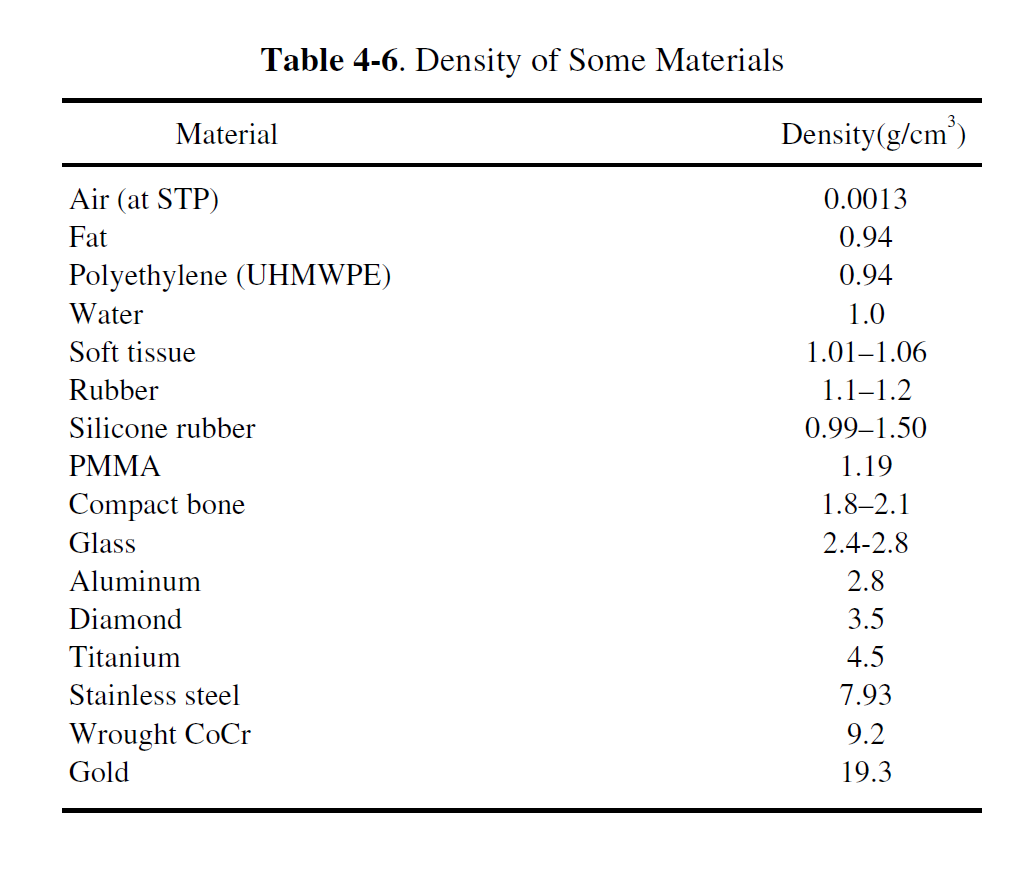}
\end{minipage}

\vspace{0.5em}
\noindent\textbf{Benchmark ID:} CAL 16 / id 024\\
\textbf{Failure type:} Wrong formula use\\
\textbf{Manual diagnosis:} The trace explicitly includes the density of air in the pores when computing apparent density and sheet weight. The reference solution uses only the solid polyurethane volume fraction, so the model applies a superficially plausible but inappropriate density formula.

\vspace{0.75em}
\noindent\textbf{Original Question}
\noindent \textbf{Example 4.5} \\ A polyurethane open cell foam is to be used for a lining for an artificial leg. The solid volume fraction is 4\%, the pores are 0.5 mm in diameter and the solid polyurethane has a density of 0.9 g/cm$^3$. Determine the porosity and apparent density of the material, and the weight of a sheet 200 mm by 200 mm by 1 cm thick.Round to 3 decimal places.

\noindent\textbf{Ground Truth Answer Slots}
[
  "0.96",
  "0.036",
  "0.032"
]

\noindent\textbf{GPT-5.5 Reasoning Trace}
Solid volume fraction = 0.04, so porosity = 1 - 0.04 = 0.96. For open-cell foam, apparent density includes solid polyurethane plus air in pores: $\rho$ = 0.04(0.9) + 0.96(0.0013) = 0.037248 g/cm³. Sheet volume = 20 cm × 20 cm × 1 cm = 400 cm³, so mass/weight = 0.037248 × 400 = 14.8992 g. Rounded to 3 decimals.

\noindent\textbf{GPT-5.5 Answer Block}
<answer>["0.960", "0.037", "14.899"]</answer>

\subsection{Case 9: Similar formula pattern but wrong Hess-law application}

\noindent
\begin{minipage}{\columnwidth}
\centering
\includegraphics[width=0.3\columnwidth]{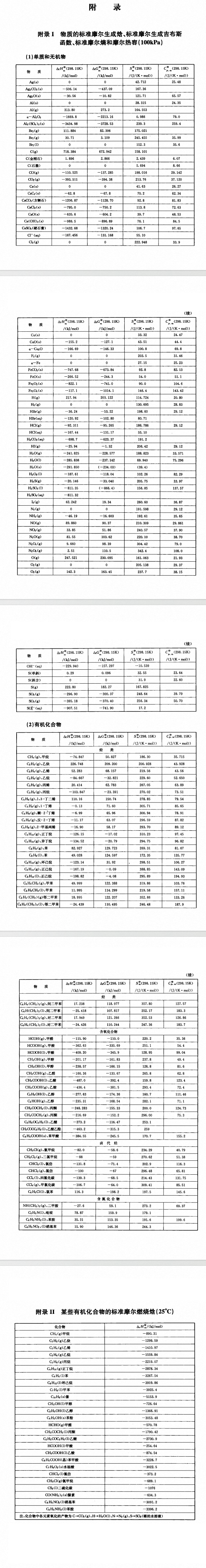}
\end{minipage}

\vspace{0.5em}
\noindent\textbf{Benchmark ID:} CAL 02 / id 062\\
\textbf{Failure type:} Similar formula, wrong application\\
\textbf{Manual diagnosis:} The model invokes the correct high-level Hess-law relation using combustion enthalpies, but applies the sign/table-value combination incorrectly. This is a useful example where the formula family is very similar to the ground-truth solution, yet the final answer is wrong.

\vspace{0.75em}
\noindent\textbf{Original Question}
Calculate the standard molar reaction enthalpy $\Delta_r H_m^\ominus$ (298 K) for the following reaction using standard molar combustion enthalpies: $CH_3COOH(l) + C_2H_5OH(l) \rightarrow CH_3COOC_2H_5(l) + H_2O(l)$. Requirements: Calculate step by step, round the result to the nearest integer, and specify the unit as kJ/mol.

\noindent\textbf{Ground Truth Answer Slots}
[
  "12"
]

\noindent\textbf{Ground Truth Reasoning Summary}
Using Hess's law with combustion enthalpies:
ΔrH°m = Σ ΔcH°m(reactants) - Σ ΔcH°m(products)
= [(-875) + (-1368)] - [(-2231) + 0] = 12 kJ/mol.

\noindent\textbf{GPT-5.5 Reasoning Trace}
Using Hess’s law with combustion enthalpies: ΔrH°m = ΔcH°m(CH3COOH,l) + ΔcH°m(C2H5OH,l) - ΔcH°m(CH3COOC2H5,l). Substituting the tabulated values gives approximately -3.9 kJ/mol, which rounds to -4 kJ/mol.

\noindent\textbf{GPT-5.5 Answer Block}
<answer>["-4 kJ/mol"]</answer>


\subsection{Case 10: Similar lever-rule formula but wrong phase-fraction assignment}

\noindent
\begin{minipage}{\columnwidth}
\centering
\includegraphics[width=0.82\columnwidth]{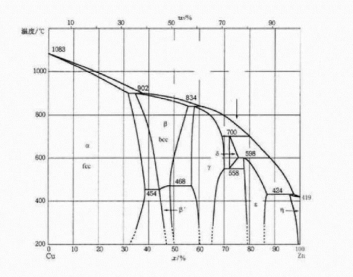}
\end{minipage}

\vspace{0.5em}
\noindent\textbf{Benchmark ID:} CAL 01 / id 054\\
\textbf{Failure type:} Similar formula, wrong parameter assignment\\
\textbf{Manual diagnosis:} The model correctly recognizes that the phase diagram calls for a lever-rule calculation, but swaps or misassigns the tie-line distances for the \(\varepsilon\) and \(\gamma\) phases.

\vspace{0.75em}
\noindent\textbf{Original Question}
Calculate the phase composition mass fraction of a Cu-75\%Zn alloy at 200$^\circ$C when cooled from equilibrium to this temperature. Present the final results as a percentage with one decimal place for each phase fraction.

\noindent\textbf{Ground Truth Answer Slots}
[
  "71.4",
  "28.6"
]

\noindent\textbf{Ground Truth Reasoning Summary}
At 200°C, the ε phase has composition 79\% Zn, the γ phase has composition 65\% Zn, and the alloy composition is 75\% Zn. Lever rule:
$f_{\epsilon}$ = (75 - 65) / (79 - 65) $\times$ 100\% = 71.4\%;
$f_{\gamma}$ = (79 - 75) / (79 - 65) $\times$ 100\% = 28.6\ \%.

\noindent\textbf{GPT-5.5 Reasoning Trace}
Using the 200°C tie-line from the Cu–Zn phase diagram, the alloy lies in the γ + ε two-phase region; applying the lever rule gives the phase mass fractions.

\noindent\textbf{GPT-5.5 Answer Block}
<answer>["30.0", "70.0"]</answer>


\subsection{Case 11: Image understanding error in recovery-kinetics plot}

\noindent
\begin{minipage}{\columnwidth}
\centering
\includegraphics[width=0.72\columnwidth]{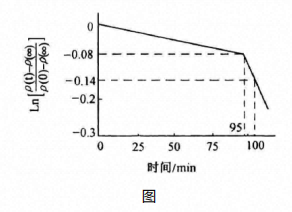}
\end{minipage}

\vspace{0.5em}
\noindent\textbf{Benchmark ID:} CAL 01 / id 077\\
\textbf{Failure type:} Image/timing interpretation error\\
\textbf{Manual diagnosis:} The model reads the second segment as a 5-minute change from \(-0.08\) to \(-0.14\), while the reference uses the 100-minute comparison at \(22^\circ\)C. This graph-reading error propagates into a much larger activation energy.

\vspace{0.75em}
\noindent\textbf{Original Question}
An aluminum sample was heated to \(284^\circ\text{C}\) and quenched to \(0^\circ\text{C}\), then immediately held at \(0^\circ\text{C}\) while the specific resistance \(\rho(t)\) was measured at different times. After holding for 95 minutes, it was immediately heated to \(22^\circ\text{C}\), held at this temperature, and the change in specific resistance with time was measured. The measured data were plotted as a curve (shown in the figure below). Using this equation and the data from the figure, determine the migration activation energy of vacancies in aluminum. Retain exactly 2 decimal places for the final result, with the unit in \(\text{J}/\text{mol}\).

\noindent\textbf{Ground Truth Answer Slots}
[
  "1569.00"
]

\noindent\textbf{Ground Truth Reasoning Summary}
From the figure: at 273 K, t = 95 min and ln ratio = -0.08; at 295 K, t = 100 min and ln ratio = -0.14. Dividing the two kinetic equations gives Q = 15.69 × $10^3$ J/mol.

\noindent\textbf{GPT-5.5 Reasoning Trace}
From the graph, the 0°C segment changes from 0 to -0.08 over 95 min, so \(k_0=0.08/95\ \text{min}^{-1}\). The 22°C segment changes from -0.08 to -0.14 over 5 min, so \(k_{22}=0.06/5\ \text{min}^{-1}\). Since \(k=Ae^{-Q/RT}\), \(\ln(k_{22}/k_0)=Q/R(1/T_0-1/T_{22})\). Using \(T_0=273\ \text{K}\), \(T_{22}=295\ \text{K}\), and \(R=8.314\ \text{J mol}^{-1}\text{K}^{-1}\), \(Q=8.314\ln[(0.06/5)/(0.08/95)]/(1/273-1/295)=80858.15\ \text{J/mol}\).

\noindent\textbf{GPT-5.5 Answer Block}
<answer>["80858.15 J/mol"]</answer>

\end{document}